\pdfoutput=1
\documentclass[11pt]{article}

\usepackage[preprint]{acl}
\usepackage{times}
\usepackage{latexsym}
\usepackage[T1]{fontenc}
\usepackage[utf8]{inputenc}
\usepackage{microtype}
\usepackage{amsmath,amssymb,mathtools,amsthm}
\usepackage{booktabs,multirow}
\usepackage{graphicx}
\usepackage{xcolor}
\usepackage{url}
\usepackage{fvextra}

\definecolor{aclblue}{HTML}{000099}
\hypersetup{colorlinks=true,linkcolor=aclblue,citecolor=aclblue,urlcolor=aclblue}

\newcommand{\silica}{\textsc{Silica}}
\newcommand{\codeurl}{\url{https://github.com/explcre/silica}}
\newcommand{\dataurl}{\url{https://huggingface.co/datasets/ryan1025-x/silica}}
\newcommand{\version}{\mathcal{V}}
\newcommand{\hypotheses}{\mathcal{H}}
\newcommand{\meanings}{\mathcal{Z}}
\newcommand{\render}{R}
\newtheorem{proposition}{Proposition}

\title{SILICA: Certified Counterfactual Evaluation of Identifiability in
Unseen-Language Induction}

\author{Pengcheng Xu \\
  University of California, Irvine \\\And
  Xinyu Guan \\
  University of Glasgow \\}

\begin{document}

\maketitle
\begingroup\renewcommand{\thefootnote}{}\footnotetext{Accepted to Findings of the Association for Computational Linguistics: EACL 2027.}\endgroup

\begin{abstract}
Knowing an answer is not the same as knowing that the evidence determines it:
several unseen lexicons or grammars can fit every clue in a prompt yet disagree
on a query.  We introduce SILICA, a shortcut-audited benchmark of certified
identifiability.  Every record publishes an executable Boolean program over a
finite lexicon--grammar class, and exhaustive filtering proves consensus or
disagreement between compatible witnesses.  Its 12,288 records span 192 packed
templates; each of 6,144 program/evidence states is queried two complementary
ways with opposite correct actions, at constant class size (128) and posterior
size (16).  Frozen shortcut controls and learned probes score exactly 50\%, and
a solver reading only the prompt reproduces all 12,288 gold labels.  Of eight
sealed silent single-pass panel conditions up to 32B, the four passing the
format gate (all Qwen2.5) answer almost independently of gold answerability with
payloads near chance, and no condition completes any of the 192 eight-row route
groups.  A 235B model permitted a visible reasoning trace escapes that regime,
reaching .671 balanced strict joint accuracy and 11/192 complete groups, while
the same checkpoint without the trace stays at $-.005$ signal and 0/192: size
alone does not move the endpoints, and 94\% of groups stay unsolved.
\end{abstract}

\section{Introduction}

Suppose two lexicons explain every translation in a prompt.  Under one,
\textit{v0} means \textsc{see}; under the other, it means \textsc{help}.  If the
query asks for \textit{v0}, producing either translation selects an arbitrary
world not licensed by the evidence.  The scientifically correct output is not
a lower-confidence guess: it is the claim that the answer is
\emph{underdetermined}.  Once a discriminating clue arrives, the same query can
become answerable.

Large language models are increasingly asked to infer rules, tool semantics,
schemas, and unfamiliar languages from context \citep[e.g.,][]{guan2026cicl}.
In each setting, a model that
can solve identifiable cases but cannot recognize non-identifiability will
confidently complete missing conventions.  Conversely, indiscriminate refusal
can look accurate whenever unanswerable examples dominate, much as a model that
ignores all context looks robust to misleading context \citep{sun2026scope}.
We therefore ask a
narrow question: \textbf{can a model distinguish what follows uniquely from
the supplied evidence from what remains one of several compatible choices, and
then execute the unique answer?}

\begin{figure*}[t]
  \centering
  \includegraphics[width=\textwidth]{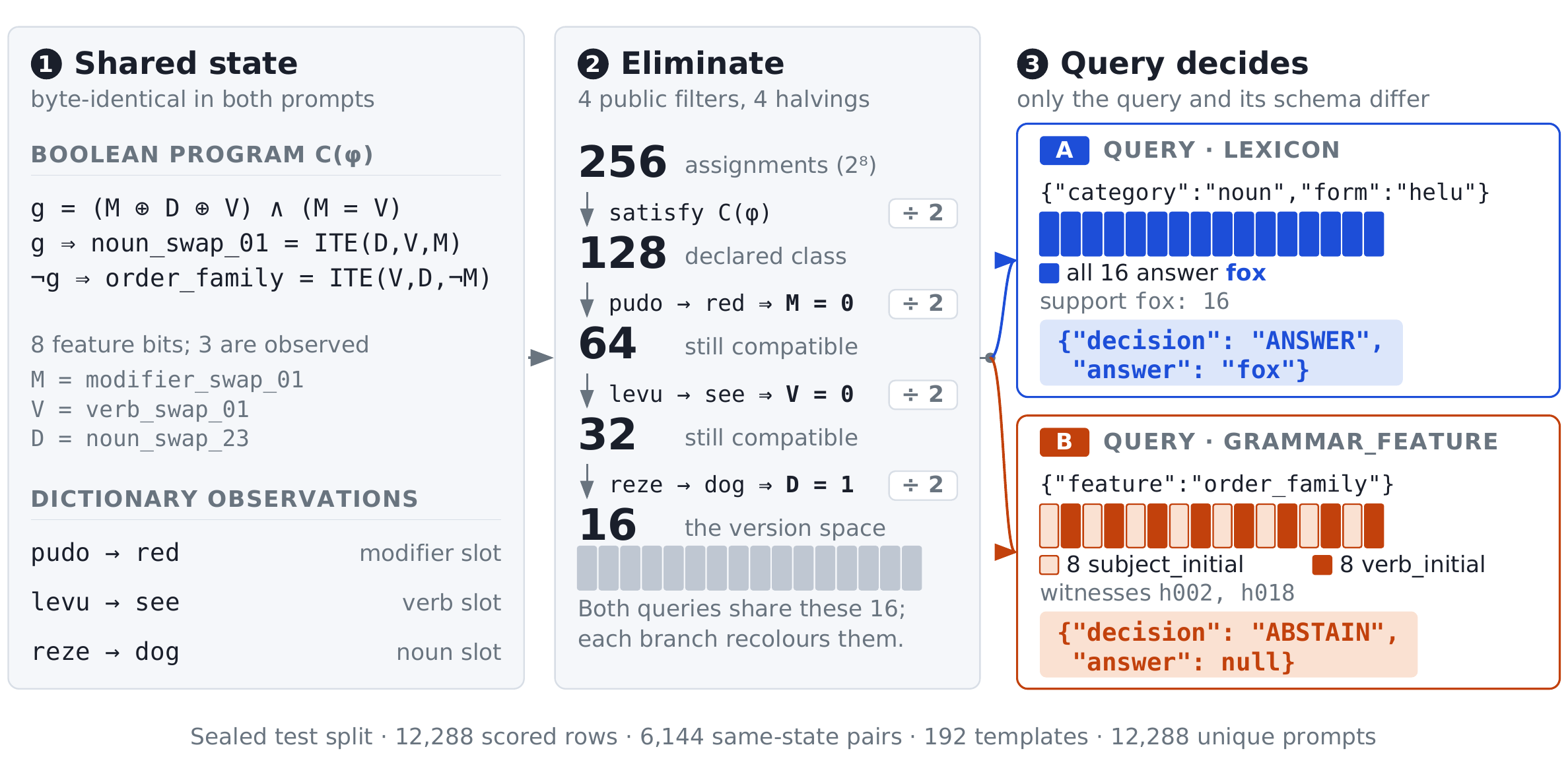}
  \caption{\textbf{One state, two opposite correct actions.}
  A renderer-faithful excerpt of one immutable released query pair;
  Appendix~\ref{app:complete-prompts} prints both complete two-message inputs
  and their prompt hashes.  The hypothesis declaration and projected
  observations are byte-identical; the renderer necessarily changes the
  structured query, \texttt{response\_task}, and task-conditioned response
  schema.  Panel~2 replays the elimination one public filter at a time,
  $256\rightarrow128\rightarrow64\rightarrow32\rightarrow16$: the program
  $C(\varphi)$ halves the assignments, then the three dictionary facts pin
  $M{=}0$, $V{=}0$ and $D{=}1$, leaving the same 16 survivors for both
  queries; every ratio shown is measured from the replayed counts.  Each
  square is one survivor in sealed certificate order, and the two branches
  recolour that same set: all 16 map \texttt{helu} to \textsc{fox}, whereas
  \texttt{order\_family} splits 8/8 with witnesses \texttt{h002} and
  \texttt{h018}, yielding the strict JSON actions \texttt{ANSWER} and
  \texttt{ABSTAIN}.  The survivors are recomputed from the public record and
  reproduce the sealed version-space hash; support counts, witnesses, and gold
  objects are evaluator-only.}
  \label{fig:main}
\end{figure*}

This capability differs from calibrated confidence.  Confidence concerns a
model's probability of correctness, and low confidence can pick which steps of
an agent's rollout to correct \citep{zhang2026uopd}; identifiability is a
property of an evidence set, query, and declared hypothesis class.  It also differs from
ordinary unanswerable question answering: \silica{} constructs a checkable
reason for non-answerability and can make the label change under a controlled
change to evidence values.  Figure~\ref{fig:main} gives the central example.

Under a silent single-pass protocol, on a sealed eight-condition panel, the
answer is no, and in one specific way: each condition answers at a rate
essentially unrelated to whether the evidence determines an answer
($|\mathrm{AR}-\mathrm{FCR}|\le.017$), so scale slides it from ``always
answer'' to ``almost always abstain'' without completing one of the 192 route
groups.  Permitting a visible reasoning trace on a much larger reasoning model
breaks that regime without solving the task, while the same permission does not
help a small reasoning model and a visible scratchpad does not help two panel
models: the benchmark is a gradient rather than a floor.

We make five contributions:
\begin{enumerate}
  \item We formalize answerability as consensus over an explicitly declared
        finite lexicon--grammar version space and provide independently
        reconstructible consensus or disagreement certificates.
  \item We construct posterior-matched, same-state complementary query pairs:
        the declaration and projected observations are unchanged, while the
        structured query, response task, and task-conditioned schema differ;
        one query must be answered and the other abstained from.
  \item We separate scored rows from semantic scale through deterministic
        distance-gated packing, 192 primary template clusters, an independent
        clean-room implementation, and exact plus learned shortcut gates.
  \item We freeze before output inspection a provenance-sealed direct-model panel and strict paired
        endpoints that separate decision orientation, exact payload execution,
        complete-group success, response validity, and false certainty, and we
        show that under a silent single-pass protocol the panel collapses onto
        two constant policies: answering carries almost no information about
        answerability, payloads stay near the exact within-unit chance level,
        and the complete-group endpoint is zero for every condition.
  \item We isolate the protocol's confounds with matched controls: a visible
        reasoning trace lifts a 235B reasoning model off the constant-policy
        frontier but not a 1.7B one, decoding is not the cause, and
        scratchpads, prompt stripping and a stated shortcut rule leave two
        panel models at chance, although the gold decision needs at most four
        Boolean operations on three printed bits.
\end{enumerate}

We deliberately do not claim broad natural-language validity, human-like
acquisition, or a scaling law.  The finite micro-language setting trades
ecological coverage for a capability label that can be proved exactly.  We
release the benchmark, its generator and validators, the evaluator, and every
model response analyzed here.\footnote{Code: \codeurl; data and model
responses: \dataurl.}

\section{Positioning and Related Work}

Human-authored linguistic puzzles test induction from few examples
\citep{sahin2020puzzling,chi2024modeling,bean2024lingoly,sanchez2024linguini},
while grammar-book and in-context translation evaluate learning from richer
resources \citep{tanzer2023mtob,aycock2025grammarbook,zhang2024teaching,
marmonier2025explicit}.  Controlled languages and formal tasks make compositional
generalization measurable \citep{lake2018scan,ruis2020gscan,kim2020cogs,
valvoda2022formal}, and synthetic impossible languages probe what language
models can learn \citep{kallini2024impossible}.  FOLIO pairs natural-language
reasoning with machine-verified first-order-logic annotations
\citep{han2024folio}, SciTaRC pairs scientific table questions with manually
constructed reasoning plans that separate planning from execution
\citep{wang2026scitarc}, whereas InductionBench tests recovery of latent
subregular functions \citep{hua2025inductionbench}.  VBVR brings rule-based,
verifiable scoring to video reasoning at scale \citep{wang2026vbvr}.  \silica{} complements
these resources: its target is neither translation quality, unrestricted
deduction, nor recovery of one intended hidden rule, but the meta-decision of
whether \emph{every declared rule still compatible with the observations}
entails the same query answer.

AmbiEnt and DUST directly study linguistic ambiguity and semantic
underspecification, respectively \citep{liu2023ambient,wildenburg2024dust};
AbstentionBench evaluates abstention across twenty datasets spanning unknown
answers, underspecification, false premises, subjective interpretations and
outdated information \citep{kirichenko2025abstentionbench}; and selective
classification formalizes prediction with a reject option.  In the noise-free
setting its consistent selective strategy abstains exactly where the version
space disagrees \citep{elyaniv2010selective}, which is the decision rule
\silica{} certifies.  \silica's distinctive supervision is therefore not the
rule but an exhaustive certificate over a declared, in-prompt finite class,
plus matched changes in evidence values.  This makes non-answerability causal within the
declared construction rather than an annotator judgment or a property inferred
from a single reference.  \silica's evidence is noise-free and consistent, so
non-answerability comes from underdetermination rather than from conflicting or
corrupted evidence, a separate failure studied for LLM agents that combine tool
outputs \citep{zhang2026conflict}.  Evidence sufficiency also shapes attributed
generation, where a single cited sentence is often not enough to ground a claim
\citep{wang2026finer}.

Appendix~\ref{app:capability-map} maps which of these design features each
neighbouring resource states as primary, and which \silica{} adds.
Benchmark artifacts can support unexpectedly strong shortcuts
\citep{gururangan2018artifacts,li2026grace}, and a failing partial-input baseline does not
certify their absence \citep{feng2019partial}.  The released construction is
therefore the third, after two candidate designs were rejected by explicit
shortcut audits.  Its same-state complementary queries are contrast-set-like
local interventions \citep{gardner2020contrast}, but the opposite labels follow
from exhaustive semantics rather than annotation.  Query-masked exact controls
and word/character probes under held-out-template and held-out-family splits
test whether the intervention survives beyond identity matching.  Calibration
methods correct biased output probabilities \citep{zhao2021calibrate} or align
confidence with correctness \citep{jiang2021calibration};
our gold answerability is instead an exactly computed property of the declared
finite problem, with false certainty reported as a behavioral consequence.

\section{Certified Identifiability}

\subsection{Executable languages and version spaces}

Let $z\in\meanings$ be a typed event meaning and let a language hypothesis
$h=(\ell,g)$ pair a part-of-speech-preserving lexicon with an executable
grammar.  Its deterministic realizer $\render_h:\meanings\rightarrow\Sigma^*$
maps meanings to synthetic strings.  Meanings are sampled before realization,
so the unknown utterance and canonical task answer are two views of one
structure rather than a word substitution.

An episode publishes a finite class $\hypotheses_e$ but hides the sampled true
member.  Dictionary observations and parallel meaning--surface pairs form
evidence $E$.  Exact filtering yields the version space
\citep{mitchell1982generalization}
\begin{equation}
\version_e(E)=\{h\in\hypotheses_e:h\models E\},
\label{eq:version-space}
\end{equation}
where $h\models E$ means that $h$ satisfies every observation in $E$.
For task $t$ and query $q$, define the supported canonical answers
\begin{equation}
A_t(q,E)=\{f_t(h,q):h\in\version_e(E)\}.
\label{eq:support}
\end{equation}
Gold is \texttt{ANSWER} with the unique member of $A_t$ when $|A_t|=1$ and
\texttt{ABSTAIN} otherwise.  The certificate stores canonical hashes of the
survivor and answer-support sets.  An underdetermined item additionally stores
two surviving hypotheses with different answers; an answerable item is
rechecked for unanimity.  None of these fields is model-visible.

\begin{proposition}[Evidence-relative necessity]
If $|A_t(q,E)|>1$, no learner observing only $(\hypotheses_e,E,q)$ can guarantee
the correct task answer for every possible true language in $\version_e(E)$.
\end{proposition}
The proof is immediate from two observationally indistinguishable witnesses
with different task answers and is given in Appendix~\ref{app:proofs}.  The
claim is deliberately relative to $\hypotheses_e$; it is not uniqueness among
all conceivable human languages.

\subsection{Adversarial construction history}

We treat shortcut discovery as part of benchmark construction.  In the initial
factorial control, a train-fit budget--query rule that ignored evidence values
reached 93.05\% balanced decision accuracy.  A first revision removed that
route but leaked every label through public posterior cardinality.  Both designs
are retained as falsified development history, not benchmark evidence.  The released
construction was accepted only after constant-cardinality, paired-query, learned-probe, and
independent clean-room gates were fixed before direct-model inference.

\subsection{Posterior-matched guarded switches}
\label{sec:stress}

The base language inventory has eight binary lexical and grammatical features,
so $\mathcal H_0=\{h_b:b\in\{0,1\}^8\}$ contains 256 executable languages.
For semantic template $\phi$, the item publishes a canonical Boolean program
$C_\phi$ and declares
\begin{equation}
\mathcal H_\phi=\{h_b\in\mathcal H_0:C_\phi(b)=1\}.
\label{eq:hard-class}
\end{equation}
Every released program admits exactly 128 languages.  Three dictionary facts
reveal a selected feature triple $X$; exact filtering leaves 16 candidates for
every evidence state.

The program implements a guarded switch between two query-facing coordinates.
Conditioned on any released $X$, exactly one coordinate is constant over all 16
survivors while the other retains both values.  We therefore ask two queries of
the same program/evidence state.  Their hypothesis declaration and projected
observations are identical.  In the rendered input, the structured query,
response task, and task-conditioned canonical response schema change: one
requires \texttt{ANSWER} with the unanimous payload, and the other requires
\texttt{ABSTAIN} with two explicit evaluator-only witnesses.

\paragraph{A worked instance.}
Take the released parity item of Figure~\ref{fig:main}.  Its published program is
$g=(M\oplus D\oplus V)\wedge(M{=}V)$, with
$g\Rightarrow\texttt{noun\_swap}_{01}=\mathrm{ITE}(D,V,M)$ and
$\neg g\Rightarrow\texttt{order\_family}=\mathrm{ITE}(V,D,\neg M)$, where
$M,V,D$ abbreviate three of the eight binary features.  Of the $2^8=256$ base
languages, exactly 128 satisfy the program.  The three dictionary facts
\texttt{pudo}$\rightarrow$\textsc{red}, \texttt{levu}$\rightarrow$\textsc{see}
and \texttt{reze}$\rightarrow$\textsc{dog} pin $M{=}0$, $V{=}0$ and $D{=}1$,
leaving 16 survivors.  All 16 map \texttt{helu} to \textsc{fox}, so the lexical
query is answerable; the same 16 split 8/8 between subject-initial and
verb-initial order, so the grammar query is not.  Nothing in this derivation is
withheld: the program, the three facts, the realization rules and the decision
rule are all in the prompt, and Figure~\ref{fig:deduction} in
Appendix~\ref{app:deduction} walks the whole elimination step by step.  Only the survivor set, its support partition and
the two disagreeing witnesses are evaluator-only.

A paired route contains four evidence states crossed with the two query roles.
The states balance active/inactive guards and both payload values, so each
four-row orientation is a complete $2\times2$ decision--payload unit and the
linked eight rows contain four examples of each action.  The public release
uses lexical and grammar-feature queries with exact canonical answers.  Four
families encode the guarded-switch principle as parity guards (implications),
majority guards (disjunctions), equality chains, and selector multiplexers
(if--then--else); they are structurally different encodings, not four claims of
unrelated linguistic mechanisms.

\subsection{Semantic packing, routes, and separation}

Scale comes from distinct behavior, not renamed formulas.  Candidate
enumeration yields 2,644 family-specific programs (2,411 globally unique paired
behaviors).  We compare programs by a length-16 ternary outcome signature: one
position for each query role at each of eight $X$ states, taking values
\texttt{ABSTAIN}, \texttt{ANSWER:0}, or \texttt{ANSWER:1}.  Candidates must
be at Hamming distance at least four positions from every fixed support
template and every selected evaluation template, both under this paired
signature and the derived full-class truth table.

The deterministic search checks all 24 family orders under two static conflict
scores.  The best balanced packing found across the 48 frozen configurations
has 57 templates per family.  We select 48 (192 total), leaving measured
headroom rather than adding nuisance variants.  Twelve support templates
are semantically disjoint from evaluation, with the selector family held out
entirely from support.  Eight lexical--grammar routes per evaluation template
cover every public $X$ state and each query axis twice.  Ordered placeholder
roles, target languages, survivors, and certificates remain evaluator-only.

\section{Evaluation}

\subsection{Prevalence-robust selective metrics}

Let $G\in\{A,U\}$ be the gold action and $\widehat G$ the predicted action.
Answerable decision recall (AR) and underdetermined abstention recall (UR) are
\begin{align}
\mathrm{AR}&=P(\widehat G=A\mid G=A),\\
\mathrm{UR}&=P(\widehat G=U\mid G=U).
\end{align}
Their unweighted mean, balanced decision accuracy
$\mathrm{BDA}=(\mathrm{AR}+\mathrm{UR})/2$, prevents a policy or budget's label
prevalence from dominating the score.  Exact answerable recall (EAR) replaces
the first event with \texttt{ANSWER} plus the canonical payload; balanced joint
accuracy is $\mathrm{BJA}=(\mathrm{EAR}+\mathrm{UR})/2$.  We also report
micro-averaged strict joint accuracy, answer coverage, invalid response rate,
and false certainty
\begin{equation}
\mathrm{FCR}=P(\widehat G=A\mid G=U).
\end{equation}
Two derived quantities separate policy from capability, because AR, UR and
$\mathrm{EAR}$ each confound them, much as end-to-end accuracy hides which
component of an agentic pipeline makes a decision
\citep{chaughule2026who}.  The \emph{decision signal}
$\mathrm{AR}-\mathrm{FCR}$ is zero exactly when a predictor's \texttt{ANSWER}
rate does not depend on the gold action, whatever that rate is; on format-valid
rows $\mathrm{BDA}=\tfrac12+\tfrac12(\mathrm{AR}-\mathrm{FCR})$, so a vanishing
decision signal forces chance-level balanced decision accuracy.  \emph{Conditional
payload accuracy} $\mathrm{CPA}=\mathrm{EAR}/\mathrm{AR}$ measures exact payload
execution among answerable rows the model actually chose to answer, and is left
undefined rather than imputed when $\mathrm{AR}=0$.  Its evidence-blind
reference is exactly $1/2$: within every four-cell unit the two answerable rows
carry different canonical payloads, so a payload chosen without reading the
evidence values is correct on exactly one of them.
Metrics are undefined rather than imputed when a stratum lacks either gold
class.  Budgets and tasks remain separate in primary tables.

Paired endpoints prevent a globally inverted predictor from looking competent.
A four-row orientation succeeds only if its entire gold decision pattern is
correct; strict success additionally requires both answerable payloads.  The
linked all-eight endpoint requires both orientations and all four same-state
query complements to be correct.  We also report unoriented complement and
factorial consistency, but label them structural diagnostics rather than
accuracy because inversion can satisfy them.

\subsection{Strict direct-model protocol}

Direct models see the response task, Boolean DSL semantics, canonical public
program, feature-value axes, realization rules, three observations, query,
decision rule, and exact response schema.  The instruction is explicit, not
open-ended: the system message names the exact procedure and the declaration
repeats the universe and filter.  No worked example is given and the model is
told to reason silently, so the panel measures single-pass execution of a stated
algorithm, not discovery of one.  They never see family, route,
template or pairing metadata, observation IDs, semantic hashes, structural
statistics, survivors, witnesses, or gold fields.  The response must be one
whole JSON object with exactly \texttt{decision} and \texttt{answer}:
\texttt{ANSWER} couples to a canonical non-null payload and \texttt{ABSTAIN}
couples to JSON null.

Each condition in the frozen original local panel runs Hugging Face Transformers
\citep{wolf2020transformers} on one GPU with bfloat16, batch size two, greedy decoding, temperature and seed zero, at most 384 new tokens,
the tokenizer chat template, and no response repair.  A generation that does
not terminate with EOS is invalid.  Any tensor-parallel extension is a
separately frozen condition.  Hosted inference is a separate stratum with
provider pinning, disabled fallbacks, JSON mode, and an independently archived
request identity.  Schema violations and hosted request failures count as
wrong; local operational faults abort the archive rather than fabricating
model outputs.

Every run binds the dataset release, public cohort, post-selection gold audit,
model and tokenizer revisions, complete weight/tokenizer file inventory,
effective generation configuration, prompt and request hashes, source commit,
executed source-file hashes, runtime libraries, GPU, token IDs, and one UUID.
The evaluator reparses raw text and recomputes these identities.  Source, cache,
weight, or configuration drift invalidates the run rather than creating a new
unlabeled condition.

\subsection{Frozen panel and controls}

The common cohort contains 1,536 rows: one complete eight-row route for each of
192 templates.  The pre-output-frozen local panel spans Qwen2.5 at 0.5B, 1.5B, 3B,
7B, 14B, and 32B \citep{yang2025qwen25}; Phi-3.5-mini
\citep{abdin2024phi3}; and SmolLM2-1.7B \citep{allal2025smollm2}, all at
immutable revisions.  Harness configuration alone can flip benchmark verdicts
\citep{li2026safetyrepro}, so the panel is fixed before any output is inspected.
Qwen2.5-14B is additionally evaluated on all 12,288 rows; Qwen2.5-32B is the
stronger common-cohort sentinel; and Qwen2.5-1.5B has
a cross-hardware replicate that is an implementation check rather than another
sample.  Hosted free models are reported only if a complete provider-pinned
1,536-row archive is obtained; canaries and partial quota-limited runs never
enter model comparisons.

Always-answer and always-abstain each attain 50\% balanced decision accuracy.
Exact generator and independent clean-room oracles must attain 100\% strict
joint accuracy, while every frozen query-masked exact control and query-visible
learned probe must remain at 50\%.  These are construction gates, not competitors.  A generic code
agent may legitimately enumerate the class and reach the oracle ceiling; that
condition measures executable access, not direct LLM reasoning or agentic
planning.

\section{Scale and Statistical Units}

\silica{} is a finite-registry benchmark with deliberately repeated views of each
logical object.  Its 12,288 scored records are organized as 1,536 paired route
instantiations of 192 semantically packed templates; the row count is therefore
coverage, not an iid sample-size claim.  Table~\ref{tab:scale} makes every
level and its statistical role explicit.

\begin{table*}[t]
\centering
\scriptsize
\caption{Exact \silica{} hierarchy.  Only the 192 semantic templates are primary
generalization clusters; lower levels support paired and complete-unit
endpoints.}
\label{tab:scale}
\begin{tabular}{@{}lrp{4.0cm}p{6.0cm}@{}}
\toprule
Level & Count & Composition & Scientific role \\
\midrule
Scored records & 12,288 & $192\times8$ routes $\times8$ rows & Repeated measurement views; never treated as iid examples \\
Complementary query pairs & 6,144 & Same program/evidence state, two queries & Query-conditioned identifiability contrast \\
Four-cell evaluator units & 3,072 & Two \texttt{ANSWER} and two \texttt{ABSTAIN} cells & Exact all-four endpoint \\
Paired route groups & 1,536 & Two linked four-cell units & Exact all-eight endpoint \\
Semantic-template clusters & 192 & Four mechanism families $\times48$ & Primary family-stratified resampling unit \\
Support templates & 12 & Fixed and semantically disjoint from evaluation & Packing/leakage gate; never scored \\
\bottomrule
\end{tabular}
\end{table*}

Figure~\ref{fig:dataset-geometry} in Appendix~\ref{app:registry} draws the same
hierarchy together with the nuisance-balance gates and the sealed packing
frontier.  Each template contributes eight lexical--grammar query-axis routes.  Each route
contains four evidence states queried on both conditionally constrained
coordinates, yielding opposite actions for every same-state pair.  Across the
full artifact this gives 1,536 unique full-class semantics, 1,536 unique paired
query behaviors, and 3,072 query tasks.  The common 1,536-row model cohort keeps
one complete four-state route per template; only the full 12,288 rows cover all
eight routes and all eight public $X$ states for every template.

The registry is selected by deterministic distance-gated packing.  Its minimum
evaluation--evaluation and evaluation--support distance is four positions on a
length-16 ternary paired-behavior signature.  At $d=4$, the balanced packing
across 48 frozen greedy configurations contains 57 templates per family; selecting 48
leaves nine templates of measured headroom.  Every row has declared class size
128 and posterior size 16, actions are exactly balanced, all frozen
query-masked exact controls and query-visible learned probes score 50\%, and
all 12,288 rendered prompts have distinct verified hashes.  An independent
clean-room oracle agrees on 12,288/12,288 records.

Model intervals use 10,000 mechanism-family-stratified resamples of the 192
templates (48 per family).  They are finite-registry cluster-sensitivity
intervals, not iid population confidence intervals or null-hypothesis tests;
exact numerators and denominators accompany percentages, and we do not interpret
one--two point differences as model rankings.

\section{Construction Audits and Model Evaluation}
\label{sec:results}

\subsection{Correctness, balance, and shortcut resistance}

Construction validity is established independently of any direct LLM-panel score.
The in-generator oracle reproduces the decision, exact payload, posterior, and
certificate on 12,288/12,288 rows.  A standard-library clean-room program that
imports no \silica{} code independently reconstructs all 1,536 programs, 204
support-plus-evaluation placeholders, route choices, semantic identities, and
distance gates; it also agrees on every row.  Reversing row order leaves all
ID-joined results unchanged, and same-seed regeneration is byte-identical.

\begin{table}[t]
\centering
\scriptsize
\caption{Release gates.  Shortcut values are exact held-out balanced
decision accuracy; no direct LLM-panel output is used.}
\label{tab:hard-audits}
\begin{tabular}{@{}p{4.0cm}r@{}}
\toprule
Gate & Audited result \\
\midrule
In-generator exact oracle & 12,288/12,288 \\
Independent clean-room oracle & 12,288/12,288 \\
Prompt-only solver (rendered input) & 12,288/12,288 \\
Declared class size 128 & 12,288/12,288 \\
Posterior size 16 & 12,288/12,288 \\
Gold action balance & 6,144/6,144 \\
Same-state opposite-action pairs & 6,144/6,144 \\
Every frozen deterministic shortcut & 6,144/12,288 \\
Word/character, held-out template/family & 6,144/12,288 each \\
Unique rendered prompt hashes & 12,288/12,288 \\
Minimum ternary-signature distance & 4 of 16 positions \\
\bottomrule
\end{tabular}
\end{table}

\paragraph{The task is decidable from the prompt alone.}
Both oracles read dataset \emph{records}, so neither answers the question a
chance-level panel provokes: does the text the model receives suffice?  We
therefore added a solver that reads only the rendered user message, importing no
oracle, certificate, or gold field.  It re-derives the feature inventory,
enumerates the declared $2^8$ universe, applies the published program, filters by
the three dictionary facts, and evaluates the query under every survivor.  It
reaches 12,288/12,288 exact strict joint accuracy, and the declared class size
128 and posterior size 16 reproduce from the prompt text on every row.  Every
scored item is therefore decidable from exactly what the model is given, so the
panel's failures are failures of reasoning rather than of missing evidence.

The strongest query-masked control omits every query-side field while keeping
the raw program and evidence, and is exactly at chance because every state
receives both labels.  Learned word and character TF-IDF logistic-regression
classifiers \citep{pedregosa2011sklearn} remove IDs,
hashes, route metadata and observed surface forms, then train under eight-fold
held-out-template and four-fold held-out-family splits; all four meet the 0.500
gate exactly.  These rule out the audited shortcut families, not every future
heuristic.

\subsection{Direct-model hardness is a measured claim}

A shortcut-resistant construction is not automatically difficult for strong
models.  We therefore froze the model identities, common/full cohorts,
generation envelope, endpoints, 10,000-replicate cluster-sensitivity procedure,
and interpretation gates before inspecting any direct-model output.
A usable run requires at least 99\% request success and 95\% strict format
validity.  We call the benchmark nonsaturated only if Qwen2.5-14B, Qwen2.5-32B,
and the cross-family Phi-3.5 configuration all have balanced strict joint
accuracy below 90\%, every upper sensitivity endpoint is below 95\%, and the
strongest usable model fails across multiple templates and mechanism families.

\begin{figure*}[t]
  \centering
  \includegraphics[width=\textwidth]{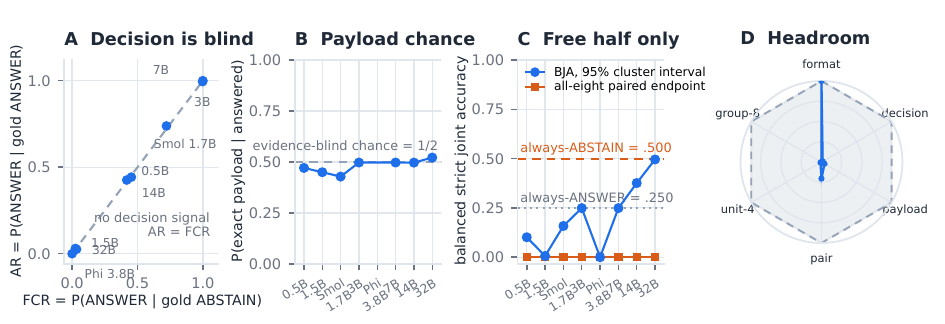}
  \caption{\textbf{Scale relocates the policy; it does not buy the capability.}
  (A) Every frozen condition lies within .017 of the
  $\mathrm{AR}=\mathrm{FCR}$ diagonal, so its choice to answer is essentially
  independent of whether the evidence determines the answer.  (B) Conditional
  on answering, exact payload accuracy stays near the within-unit chance level
  $1/2$.  (C) Across the four format-valid conditions, balanced strict joint
  accuracy climbs .249$\rightarrow$.495 only by moving between the always-\texttt{ANSWER}
  (.250) and always-\texttt{ABSTAIN} (.500) constants, while the all-eight paired
  endpoint stays at $0/192$ everywhere.  (D) Per scored endpoint, the best value
  reached by \emph{any} frozen condition, against the exact oracle at 1 (dashed).
  Response formatting is solved; the five endpoints that require using the
  evidence are almost entirely unclaimed.  Taking a per-axis maximum flatters the
  panel, so the unclaimed area is unclaimed by every model at once.  Intervals
  are 10,000-replicate family-stratified template-cluster resamples; exact
  numbers are in Appendix~\ref{app:model-panel}.}
  \label{fig:model-panel}
\end{figure*}

\paragraph{The decision carries almost no information about answerability.}
Figure~\ref{fig:model-panel} plots the panel; Table~\ref{tab:model-panel} in
Appendix~\ref{app:model-panel} prints it exactly.  Across all eight conditions
the signed gap $\mathrm{AR}-\mathrm{FCR}$ stays inside $[-.010,+.017]$
(Figure~\ref{fig:model-panel}A).  A policy answering at a fixed rate without
consulting gold satisfies $\mathrm{AR}=\mathrm{FCR}$ identically, and
$\mathrm{BDA}=\tfrac12+\tfrac12(\mathrm{AR}-\mathrm{FCR})$ on format-valid
rows; the four format-valid conditions lie within .010 of that family, and no
condition exceeds .503 balanced decision accuracy.

\paragraph{Payload execution stays near chance.}
All 3,072 four-cell units carry two \emph{different} canonical payloads on their
answerable rows, so an evidence-blind payload scores exactly $1/2$ per unit.
Conditional payload accuracy spans .429--.522, and .497--.522 for the four
format-valid conditions (Figure~\ref{fig:model-panel}B).

\paragraph{Scale relocates the policy.}
Balanced strict joint accuracy rises from .249 at 3B and 7B to .495 at 32B, and
the rise is entirely the abstention half: Qwen2.5-7B answers every row
($\mathrm{AR}=1.000$, $\mathrm{UR}=.000$) while Qwen2.5-32B abstains on almost
every row ($\mathrm{AR}=.030$, $\mathrm{UR}=.975$).  Since always-\texttt{ABSTAIN}
attains $.500$ and always-\texttt{ANSWER} with an evidence-blind payload attains
$.250$, the four format-valid conditions lie between two constants, and the strongest interval
$[.491,.499]$ falls \emph{below} the always-\texttt{ABSTAIN} constant
(Figure~\ref{fig:model-panel}C).

\paragraph{Same-state pairs are unsolved.}
The endpoint the benchmark was built for asks for opposite, correct actions on
the two queries of one state.  On this same-state pair endpoint every panel
condition except Qwen2.5-32B scores at or below the coin level $\alpha(1-\alpha)$
implied by its own answer rate $\alpha$ (32B: $23/768=.030$ against $.027$), the
all-eight endpoint is $0/192$ throughout, and Qwen2.5-14B completes the only four-cell
patterns at all ($5/384$), under the $.061$ expected from guessing.

\providecommand{\MatchedControlRows}{}
\renewcommand{\MatchedControlRows}{%
Qwen3-235B-A22B & trace off & 1.000 & $-$.005 & .489 & .408 & 0/192 & 13 \\
 & trace on & 1.000 & $+$.431 & .878 & .671 & 11/192 & 5{,}427 \\
Qwen3-1.7B & trace off & .987 & $-$.001 & .449 & .224 & 0/192 & 19 \\
 & trace on & .950 & $+$.029 & .447 & .279 & 0/192 & 3{,}868 \\
Qwen3.5-2B & greedy & 1.000 & $+$.000 & .487 & .243 & 0/192 & 20 \\
 & stochastic & 1.000 & $+$.000 & .435 & .217 & 0/192 & 18 \\
\midrule
Qwen2.5-14B & silent$^{\ast}$ & .990 & $+$.004 & .503 & .382 & 0/192 & 18 \\
 & scratchpad & .984 & $-$.007 & .496 & .307 & 0/192 & 697 \\
Qwen2.5-32B & silent$^{\ast}$ & 1.000 & $+$.003 & .526 & .495 & 0/192 & 14 \\
 & scratchpad & .961 & $+$.009 & .485 & .264 & 0/192 & 832 \\
}

\begin{table}[tb]
\centering
\small
\caption{Matched controls on the 1,536-row common cohort, final JSON scored.
Each pair changes one factor on a fixed checkpoint: the reasoning trace, the
decoding, or (panel models) a visible \texttt{<think>} scratchpad with a
4{,}096-token budget; changing two factors at once would confound them
\citep{sun2026lora}.  $^{\ast}$Released prompt served through the ablation
harness, which reproduces each sealed panel row to within .023.
$\langle 8\rangle$ counts complete eight-row route groups; Tok.\ is the mean
number of generated tokens per row.}
\label{tab:matched-controls}
\resizebox{\columnwidth}{!}{%
\begin{tabular}{@{}llrrrrrr@{}}
\toprule
Model & Condition & Valid & AR$-$FCR & CPA & BJA & $\langle 8\rangle$ & Tok. \\
\midrule
\MatchedControlRows
\bottomrule
\end{tabular}}
\end{table}

\paragraph{A visible reasoning trace escapes the constant-policy regime.}
The panel above forbids a reasoning trace.  We therefore ran one separately
frozen extension condition that permits it: Qwen3-235B-A22B
\citep{yang2025qwen3} in its GPTQ-Int4 release \citep{frantar2023optq}, served
by vLLM \citep{kwon2023vllm} with tensor-parallel 4 and thinking enabled, a 16{,}384-token budget, a reasoning
parser that separates the trace from the scored content, and scoring on the
final content channel only.  It returns a response on all 1,536 rows, with request
success and strict format validity both 1.000, and it does not lie on the constant-policy
frontier: decision signal $.431$ against $[-.010,+.017]$ for the whole panel,
conditional payload accuracy $.878$ against a chance level of $1/2$, balanced
strict joint accuracy $.671\,[.645,.697]$ against a panel maximum of $.495$,
same-state pair decision accuracy $.518$ (independent answering at its own rates
would give $.512$), and $11/192$ complete route groups where every
panel condition scores $0/192$.

The endpoints move, so they measure something achievable, and $181/192$ route
groups remain unsolved.

\paragraph{Neither size nor the trace alone moves the endpoints.}
That condition differs from the panel in three ways at once, so we ran matched
controls (Appendix~\ref{app:protocol}).  Holding the checkpoint and sampling
fixed and disabling the trace, the same 235B returns to the frontier: decision
signal $-.005$, conditional payload accuracy $.489$, and $0/192$ complete route
groups, at a mean $13$ tokens per row against $5{,}427$.  Its stochastic sampling
is thus not the cause, and on Qwen3.5-2B \citep{qwen2026qwen35} greedy and
stochastic decoding give the same always-\texttt{ANSWER} policy
(Table~\ref{tab:matched-controls}).  Permission alone is also
insufficient: Qwen3-1.7B \citep{yang2025qwen3} writes $3{,}868$ tokens per row under the same budget
yet reaches $+.029$ signal, payload accuracy $.447$ \emph{below} the chance level
of one half, and $0/192$.  At the small end, models often fail to \emph{close} the
trace at all: generation never terminates on $28.8\%$ of rows for Qwen3.5-0.8B and
$16.5\%$ for Qwen3.5-2B, against $4.4\%$ for Qwen3-1.7B and $0.0\%$ for
Qwen3-235B, with strict format validity $32.8\%$, $48.5\%$, $95.0\%$ and $100\%$.
Non-termination falls with size within each family but not across them, so we
read no scaling curve from four points.

\paragraph{Prompt interventions relocate the policy too.}
Four renderings served to Qwen2.5-14B and Qwen2.5-32B---released, a
\texttt{<think>} scratchpad at 4{,}096 tokens, a user message 25.3\% shorter on
average, and the guarded-switch route in place of the enumeration---move AR
from $.085$ to $.730$ (14B) and from $.010$ to $.845$ (32B), while the decision
signal stays in $[-.007,+.013]$ and the all-eight endpoint at $0/192$
(Appendix~\ref{app:prompt-ablations}).  The scratchpads write $697$ and $832$
tokens per row yet raise AR and FCR together.  The gold decision itself is a
Boolean function of the three revealed bits with minimal formula size at most
four (Appendix~\ref{app:guard-structure}), so on the decision axis the collapse
is not a shortage of the computation the task requires.

\paragraph{Prespecified verdict.}
Qwen2.5-14B and Qwen2.5-32B satisfy every frozen nonsaturation criterion: request
success $\geq .99$, strict format validity $\geq .95$, balanced strict joint
accuracy below .90, upper sensitivity endpoint below .95, and failures spanning
all four mechanism families and all 192 templates.  The prespecified full-artifact
sentinel has since finalized with zero errors, and it reproduces every
common-cohort finding at eight times the scale: over all 12,288 rows Qwen2.5-14B
has decision signal $-.001$, conditional payload accuracy $.502$, balanced
decision accuracy $.495$, balanced strict joint accuracy
$.379\,[.372,.385]$, and $0/1{,}536$ complete route groups.  The third prespecified hardness
model, Phi-3.5, yields no usable evidence: under our single-EOS protocol every
response carries its end-of-turn token inside the scored span, typically after
a JSON object and in 1,021 of 1,536 rows followed by a further turn.  Stopping on
its declared end-of-turn tokens removes that artifact, yet only $49/314$ rows are
format-valid by the time limit (a complete run would stay below $.83$), as $60\%$
loop to the token cap.  The verdict therefore rests on its two Qwen2.5 legs; the
third is unevaluable, not evidence of hardness.  The cross-hardware Qwen2.5-1.5B replicate reproduces
its primary-node counterpart to the last digit, an implementation check rather
than an additional sample.

\section{Discussion}

\paragraph{What the benchmark isolates.}
\silica{} makes ``knowing when to answer'' executable: retain every compatible
explanation, answer only when their projections agree, then execute the payload.
The silent panel fails at the first step, picking between the two decisions without
consulting the property that determines the answer, so its decision axis
degenerates to a constant policy whose only free parameter is how often to
answer.  This targets convention learning in schema induction, API discovery
and rule following, where silently filling an unobserved convention fails.
The same decision arises when an LLM optimizer infers from scored measurements
how coupled settings must be configured together
\citep{esmat2026ann,zhao2026nominate}, when a cloud controller commits to a
resource-risk split only if its forecast-error history supports it
\citep{xu2026sla}, and when an agent decides whether a remembered premise is
safe to act on or a claim is supported enough to enter shared memory
\citep{wang2026memory,li2026cpb}.

\section{Conclusion}

Unseen-language induction should not reward arbitrary compatible worlds.
\silica{} makes the missing step executable: certificates define entailment,
and posterior-matched queries test whether the query is used.  Silent models
trade one constant policy for another; among our conditions, only the 235B
reasoning model with a visible trace escapes.

\section{Limitations}

\silica{} is a closed synthetic instrument.  It certifies identifiability only
relative to each declared finite class, uses generated inventories and a
simplified grammar, and its four mechanism families encode one guarded-switch
principle, so it supports no claim about open-domain underspecification, tool
discovery or natural low-resource translation.  The constant-policy result
concerns format-valid, direct, silent single-pass conditions: the format-valid
evidence comes from one model family, a visible reasoning trace changes the
result for at least one large reasoning model, and the trace-permitted
conditions are exploratory controls rather than a scaling study.  Format
failures at small scale leave some conditions uninformative, the single-EOS
protocol made one prespecified model unevaluable, the shortcut audits cover
the families we specified, the prompt relies on two implicit naming
conventions, and the intervals condition on one finite registry.
Appendix~\ref{app:caveats} details each point.

\section{Ethical Considerations}

All forms and meanings are generated: there are no personal data, human-subject
annotations, copyrighted corpus passages, or speaker-community claims.  This
lowers collection risk but supports no ecological claim about real low-resource
or endangered languages.  Abstention is not automatically safe: error costs are
context-dependent, and real translation requires community governance and
qualified human review.

Generative-AI tools aided implementation, testing, literature discovery, and
drafting, but not the authoritative meanings, surfaces, labels, or certificates,
which the deterministic pipeline and validators produce.  The authors remain
responsible for all code, claims, citations, and prose.

\clearpage
\bibliography{references}

\begin{thebibliography}{58}
\providecommand{\natexlab}[1]{#1}

\bibitem[{Abdin et~al.(2024)Abdin, Aneja, Awadalla, Awadallah, Awan, Bach,
  Bahree, Bakhtiari, Bao, Behl, Benhaim, Bilenko, Bjorck, Bubeck, Cai, Cai,
  Chaudhary, Chen, Chen, Chen, Chen, Chen, Cheng, Chopra, Dai, Dixon, Eldan,
  Fragoso, Gao, Gao, Gao, Garg, Del~Giorno, Goswami, Gunasekar, Haider, Hao,
  Hewett, Hu, Huynh, Iter, Jacobs, Javaheripi, Jin, Karampatziakis, Kauffmann,
  Khademi, Kim, Kim, Kurilenko, Lee, Lee, Li, Li, Liang, Liden, Lin, Lin, Liu,
  Liu, Liu, Liu, Liu, Luo, Madan, Mahmoudzadeh, Majercak, Mazzola, Mendes,
  Mitra, Modi, Nguyen, Norick, Patra, Perez-Becker, Portet, Pryzant, Qin,
  Radmilac, Ren, de~Rosa, Rosset, Roy, Ruwase, Saarikivi, Saied, Salim,
  Santacroce, Shah, Shang, Sharma, Shen, Shukla, Song, Tanaka, Tupini,
  Vaddamanu, Wang, Wang, Wang, Wang, Wang, Wang, Ward, Wen, Witte, Wu, Wu,
  Wyatt, Xiao, Xu, Xu, Xu, Xue, Yadav, Yang, Yang, Yang, Yang, Yu, Yuan, Zhang,
  Zhang, Zhang, Zhang, Zhang, Zhang, Zhang, and Zhou}]{abdin2024phi3}
Marah Abdin, Jyoti Aneja, Hany Awadalla, Ahmed Awadallah, Ammar~Ahmad Awan,
  Nguyen Bach, Amit Bahree, Arash Bakhtiari, Jianmin Bao, Harkirat Behl, Alon
  Benhaim, Misha Bilenko, Johan Bjorck, S{\'e}bastien Bubeck, Martin Cai, Qin
  Cai, Vishrav Chaudhary, Dong Chen, Dongdong Chen, and 110 others. 2024.
\newblock \href {https://doi.org/10.48550/arXiv.2404.14219} {{Phi-3} technical
  report: A highly capable language model locally on your phone}.
\newblock \emph{arXiv preprint arXiv:2404.14219}.

\bibitem[{Aycock et~al.(2025)Aycock, Stap, Wu, Monz, and
  Sima{'}an}]{aycock2025grammarbook}
Seth Aycock, David Stap, Di~Wu, Christof Monz, and Khalil Sima{'}an. 2025.
\newblock \href
  {https://proceedings.iclr.cc/paper_files/paper/2025/hash/20f44da80080d76bbc35bca0027f14e6-Abstract-Conference.html}
  {Can {LLM}s really learn to translate a low-resource language from one
  grammar book?}
\newblock In \emph{International Conference on Learning Representations},
  volume 2025, pages 12334--12357.

\bibitem[{Bean et~al.(2024)Bean, Hellsten, Mayne, Magomere, Chi, Chi, Hale, and
  Kirk}]{bean2024lingoly}
Andrew Bean, Simi Hellsten, Harry Mayne, Jabez Magomere, Ethan~A. Chi, Ryan
  Chi, Scott~A. Hale, and Hannah~Rose Kirk. 2024.
\newblock \href {https://doi.org/10.52202/079017-0825} {{LINGOLY}: A benchmark
  of olympiad-level linguistic reasoning puzzles in low resource and extinct
  languages}.
\newblock In \emph{Advances in Neural Information Processing Systems},
  volume~37, pages 26224--26237. Curran Associates, Inc.

\bibitem[{Ben~Allal et~al.(2025)Ben~Allal, Lozhkov, Bakouch,
  Mart{\'i}n~Bl{\'a}zquez, Penedo, Tunstall, Marafioti, Piqueres~Lajar{\'i}n,
  Kydl{\'i}{\v{c}}ek, Srivastav, Lochner, Fahlgren, Nguyen, Burtenshaw,
  Fourrier, Zhao, Larcher, Morlon, Zakka, Raffel, von Werra, and
  Wolf}]{allal2025smollm2}
Loubna Ben~Allal, Anton Lozhkov, Elie Bakouch, Gabriel Mart{\'i}n~Bl{\'a}zquez,
  Guilherme Penedo, Lewis Tunstall, Andr{\'e}s Marafioti, Agust{\'i}n
  Piqueres~Lajar{\'i}n, Hynek Kydl{\'i}{\v{c}}ek, Vaibhav Srivastav, Joshua
  Lochner, Caleb Fahlgren, Xuan-Son Nguyen, Ben Burtenshaw, Cl{\'e}mentine
  Fourrier, Haojun Zhao, Hugo Larcher, Mathieu Morlon, Cyril Zakka, and 3
  others. 2025.
\newblock \href {https://openreview.net/forum?id=3JiCl2A14H} {{SmolLM2}: When
  smol goes big---data-centric training of a fully open small language model}.
\newblock In \emph{Second Conference on Language Modeling (COLM)}.

\bibitem[{Chaughule et~al.(2026)Chaughule, Zhang, Tajwar, Menta, Vats, Jain,
  Chadha, and Chaudhary}]{chaughule2026who}
Vaishnavi Chaughule, Yuchen Zhang, Mohammed~Ahnaf Tajwar, Sai~Akshay Menta,
  Saanidhya Vats, Vinija Jain, Aman Chadha, and Divya Chaudhary. 2026.
\newblock \href {https://arxiv.org/abs/2611.29829} {Who makes the decision?
  {A}uditing an agentic {LLM} pipeline for illicit cryptocurrency transaction
  monitoring}.
\newblock \emph{arXiv preprint arXiv:2611.29829}.

\bibitem[{Chi et~al.(2024)Chi, Malchev, Kong, Chi, Huang, Chi, McCoy, and
  Radev}]{chi2024modeling}
Nathan~A. Chi, Teodor Malchev, Riley Kong, Ryan~A. Chi, Lucas Huang, Ethan~A.
  Chi, R.~Thomas McCoy, and Dragomir Radev. 2024.
\newblock \href {https://doi.org/10.18653/v1/2024.sigtyp-1.14} {{M}ode{L}ing: A
  novel dataset for testing linguistic reasoning in language models}.
\newblock In \emph{Proceedings of the 6th Workshop on Research in Computational
  Linguistic Typology and Multilingual NLP}, pages 113--119.

\bibitem[{El-Yaniv and Wiener(2010)}]{elyaniv2010selective}
Ran El-Yaniv and Yair Wiener. 2010.
\newblock \href {https://jmlr.org/papers/v11/el-yaniv10a.html} {On the
  foundations of noise-free selective classification}.
\newblock \emph{Journal of Machine Learning Research}, 11(53):1605--1641.

\bibitem[{Esmat et~al.(2026)Esmat, Lacewell, Gobriel, Jain, and
  Jannesari}]{esmat2026ann}
Shahrzad Esmat, Chaunt{\'e}~W. Lacewell, Sameh Gobriel, Nilesh Jain, and Ali
  Jannesari. 2026.
\newblock \href {https://arxiv.org/abs/2606.05489} {{LLM}-guided {ANN} index
  optimization for human-object interaction retrieval}.
\newblock \emph{arXiv preprint arXiv:2606.05489}.

\bibitem[{Feng et~al.(2019)Feng, Wallace, and Boyd-Graber}]{feng2019partial}
Shi Feng, Eric Wallace, and Jordan Boyd-Graber. 2019.
\newblock \href {https://doi.org/10.18653/v1/P19-1554} {Misleading failures of
  partial-input baselines}.
\newblock In \emph{Proceedings of the 57th Annual Meeting of the Association
  for Computational Linguistics}, pages 5533--5538.

\bibitem[{Frantar et~al.(2023)Frantar, Ashkboos, Hoefler, and
  Alistarh}]{frantar2023optq}
Elias Frantar, Saleh Ashkboos, Torsten Hoefler, and Dan Alistarh. 2023.
\newblock \href {https://openreview.net/forum?id=tcbBPnfwxS} {{OPTQ}: Accurate
  quantization for generative pre-trained transformers}.
\newblock In \emph{The Eleventh International Conference on Learning
  Representations}.
\newblock Preprint version titled ``{GPTQ}: Accurate Post-Training Quantization
  for Generative Pre-trained Transformers'', arXiv:2210.17323.

\bibitem[{Friedman(2001)}]{friedman2001greedy}
Jerome~H. Friedman. 2001.
\newblock \href {https://doi.org/10.1214/aos/1013203451} {Greedy function
  approximation: A gradient boosting machine}.
\newblock \emph{The Annals of Statistics}, 29(5):1189--1232.

\bibitem[{Gardner et~al.(2020)Gardner, Artzi, Basmov, Berant, Bogin, Chen,
  Dasigi, Dua, Elazar, Gottumukkala, Gupta, Hajishirzi, Ilharco, Khashabi, Lin,
  Liu, Liu, Mulcaire, Ning, Singh, Smith, Subramanian, Tsarfaty, Wallace,
  Zhang, and Zhou}]{gardner2020contrast}
Matt Gardner, Yoav Artzi, Victoria Basmov, Jonathan Berant, Ben Bogin, Sihao
  Chen, Pradeep Dasigi, Dheeru Dua, Yanai Elazar, Ananth Gottumukkala, Nitish
  Gupta, Hannaneh Hajishirzi, Gabriel Ilharco, Daniel Khashabi, Kevin Lin,
  Jiangming Liu, Nelson~F. Liu, Phoebe Mulcaire, Qiang Ning, and 7 others.
  2020.
\newblock \href {https://doi.org/10.18653/v1/2020.findings-emnlp.117}
  {Evaluating models' local decision boundaries via contrast sets}.
\newblock In \emph{Findings of the Association for Computational Linguistics:
  EMNLP 2020}, pages 1307--1323. Association for Computational Linguistics.

\bibitem[{Guan et~al.(2027)Guan, Zhao, and Deng}]{guan2026cicl}
Xinyu Guan, Qianyang Zhao, and Yuming Deng. 2027.
\newblock \href {https://doi.org/10.1007/978-981-92-8140-4_14} {Decision-aware
  memory cards: Counterfactual-inspired context selection and compression for
  tool-using {LLM} agents}.
\newblock In \emph{Neural Information Processing (ICONIP 2026)}, volume 3202 of
  \emph{Communications in Computer and Information Science}, pages 186--200.
  Springer Nature Singapore.

\bibitem[{Gururangan et~al.(2018)Gururangan, Swayamdipta, Levy, Schwartz,
  Bowman, and Smith}]{gururangan2018artifacts}
Suchin Gururangan, Swabha Swayamdipta, Omer Levy, Roy Schwartz, Samuel~R.
  Bowman, and Noah~A. Smith. 2018.
\newblock \href {https://doi.org/10.18653/v1/N18-2017} {Annotation artifacts in
  natural language inference data}.
\newblock In \emph{Proceedings of the 2018 Conference of the North {A}merican
  Chapter of the Association for Computational Linguistics: Human Language
  Technologies, Volume 2 (Short Papers)}, pages 107--112.

\bibitem[{Han et~al.(2024)Han, Schoelkopf, Zhao, Qi, Riddell, Zhou, Coady,
  Peng, Qiao, Benson, Sun, Wardle-Solano, Szab{\'o}, Zubova, Burtell, Fan, Liu,
  Wong, Sailor, Ni, Nan, Kasai, Yu, Zhang, Fabbri, Kryscinski, Yavuz, Liu, Lin,
  Joty, Zhou, Xiong, Ying, Cohan, and Radev}]{han2024folio}
Simeng Han, Hailey Schoelkopf, Yilun Zhao, Zhenting Qi, Martin Riddell, Wenfei
  Zhou, James Coady, David Peng, Yujie Qiao, Luke Benson, Lucy Sun, Alexander
  Wardle-Solano, Hannah Szab{\'o}, Ekaterina Zubova, Matthew Burtell, Jonathan
  Fan, Yixin Liu, Brian Wong, Malcolm Sailor, and 16 others. 2024.
\newblock \href {https://doi.org/10.18653/v1/2024.emnlp-main.1229} {{FOLIO}:
  Natural language reasoning with first-order logic}.
\newblock In \emph{Proceedings of the 2024 Conference on Empirical Methods in
  Natural Language Processing}, pages 22017--22031.

\bibitem[{Haque et~al.(2025)Haque, Yang, Yang, and Xu}]{haque2025quantization}
Md~Nazmul Haque, Hua Yang, Zhou Yang, and Bowen Xu. 2025.
\newblock \href {https://arxiv.org/abs/2508.00128} {How quantization impacts
  privacy risk on {LLMs} for code?}
\newblock \emph{arXiv preprint arXiv:2508.00128}.

\bibitem[{Hua et~al.(2025)Hua, Wong, Sun, Pan, Jardine, and
  Wang}]{hua2025inductionbench}
Wenyue Hua, Tyler Wong, Fei Sun, Liangming Pan, Adam Jardine, and William~Yang
  Wang. 2025.
\newblock \href {https://doi.org/10.18653/v1/2025.acl-long.1287}
  {{I}nduction{B}ench: {LLM}s fail in the simplest complexity class}.
\newblock In \emph{Proceedings of the 63rd Annual Meeting of the Association
  for Computational Linguistics (Volume 1: Long Papers)}, pages 26526--26546.

\bibitem[{Huang et~al.(2026)Huang, Xu, Liu, Zheng, Zhou, Qi, Yao, Liu, Xu, Xu,
  Huang, Fu, and Lin}]{huang2026guide}
Youling Huang, Tiankuo Xu, Jiaji Liu, Tong Zheng, Shuo Zhou, Shaotong Qi,
  Junchi Yao, Shiyang Liu, Hao Xu, Pengcheng Xu, Bo~Huang, Hongyi Fu, and Lin
  Lin. 2026.
\newblock \href {https://arxiv.org/abs/2609.37898} {Guide, then let go:
  Gap-adaptive teacher scheduling for sparse-reward agentic {RL}}.
\newblock \emph{arXiv preprint arXiv:2609.37898}.

\bibitem[{Jiang et~al.(2021)Jiang, Araki, Ding, and
  Neubig}]{jiang2021calibration}
Zhengbao Jiang, Jun Araki, Haibo Ding, and Graham Neubig. 2021.
\newblock \href {https://doi.org/10.1162/tacl_a_00407} {How can we know when
  language models know? {O}n the calibration of language models for question
  answering}.
\newblock \emph{Transactions of the Association for Computational Linguistics},
  9:962--977.

\bibitem[{Kallini et~al.(2024)Kallini, Papadimitriou, Futrell, Mahowald, and
  Potts}]{kallini2024impossible}
Julie Kallini, Isabel Papadimitriou, Richard Futrell, Kyle Mahowald, and
  Christopher Potts. 2024.
\newblock \href {https://doi.org/10.18653/v1/2024.acl-long.787} {Mission:
  Impossible language models}.
\newblock In \emph{Proceedings of the 62nd Annual Meeting of the Association
  for Computational Linguistics (Volume 1: Long Papers)}, pages 14691--14714.

\bibitem[{Kim and Linzen(2020)}]{kim2020cogs}
Najoung Kim and Tal Linzen. 2020.
\newblock \href {https://doi.org/10.18653/v1/2020.emnlp-main.731} {{COGS}: A
  compositional generalization challenge based on semantic interpretation}.
\newblock In \emph{Proceedings of the 2020 Conference on Empirical Methods in
  Natural Language Processing (EMNLP)}, pages 9087--9105.

\bibitem[{Kirichenko et~al.(2025)Kirichenko, Ibrahim, Chaudhuri, and
  Bell}]{kirichenko2025abstentionbench}
Polina Kirichenko, Mark Ibrahim, Kamalika Chaudhuri, and Samuel~J. Bell. 2025.
\newblock \href {https://doi.org/10.52202/085713-5729} {{AbstentionBench}:
  Reasoning {LLM}s fail on unanswerable questions}.
\newblock In \emph{Advances in Neural Information Processing Systems},
  volume~38, pages 190800--190850. Curran Associates, Inc.
\newblock Datasets and Benchmarks Track.

\bibitem[{Kwon et~al.(2023)Kwon, Li, Zhuang, Sheng, Zheng, Yu, Gonzalez, Zhang,
  and Stoica}]{kwon2023vllm}
Woosuk Kwon, Zhuohan Li, Siyuan Zhuang, Ying Sheng, Lianmin Zheng, Cody~Hao Yu,
  Joseph~E. Gonzalez, Hao Zhang, and Ion Stoica. 2023.
\newblock \href {https://doi.org/10.1145/3600006.3613165} {Efficient memory
  management for large language model serving with {PagedAttention}}.
\newblock In \emph{Proceedings of the 29th Symposium on Operating Systems
  Principles}, SOSP '23, pages 611--626. Association for Computing Machinery.

\bibitem[{Lake and Baroni(2018)}]{lake2018scan}
Brenden Lake and Marco Baroni. 2018.
\newblock \href {https://proceedings.mlr.press/v80/lake18a.html}
  {Generalization without systematicity: On the compositional skills of
  sequence-to-sequence recurrent networks}.
\newblock In \emph{Proceedings of the 35th International Conference on Machine
  Learning}, volume~80 of \emph{Proceedings of Machine Learning Research},
  pages 2873--2882. PMLR.

\bibitem[{Li et~al.(2023)Li, Donatelli, Koller, Linzen, Yao, and
  Kim}]{li2023slog}
Bingzhi Li, Lucia Donatelli, Alexander Koller, Tal Linzen, Yuekun Yao, and
  Najoung Kim. 2023.
\newblock \href {https://doi.org/10.18653/v1/2023.emnlp-main.194} {{SLOG}: A
  structural generalization benchmark for semantic parsing}.
\newblock In \emph{Proceedings of the 2023 Conference on Empirical Methods in
  Natural Language Processing}, pages 3213--3232.

\bibitem[{Li et~al.(2026{\natexlab{a}})Li, Zeng, Luo, Jia, Li, Geng, Kolter,
  Lee, and Li}]{li2026witness}
Siyuan Li, Haoxuan Zeng, Xin Luo, Fernando Jia, Florence Li, Zhengyang Geng,
  J.~Zico Kolter, Tai~Sing Lee, and Tianqin Li. 2026{\natexlab{a}}.
\newblock \href {https://openreview.net/forum?id=lc1XaBdx6q} {Witness overlap:
  Directional provenance inside open-weight model families}.
\newblock In \emph{Advances in Neural Information Processing Systems}.

\bibitem[{Li et~al.(2026{\natexlab{b}})Li, Wang, Zhu, Xu, Yang, Sun, Song,
  Duan, and Cai}]{li2026cpb}
Xiaoyang Li, Yiqi Wang, Chencheng Zhu, Ke~Xu, Wencheng Yang, Zequn Sun, Pingan
  Song, Yiqun Duan, and Taotao Cai. 2026{\natexlab{b}}.
\newblock \href {https://arxiv.org/abs/2609.30813} {A benchmark and diagnostic
  study of epistemic admission in shared agent memory}.
\newblock \emph{arXiv preprint arXiv:2609.30813}.

\bibitem[{Li et~al.(2027)Li, Xia, Zhao, Xu, Liu, Lu, Long, Ma, Cao, Fan, and
  Xu}]{li2026grace}
Xinjin Li, Yudi Xia, Xi~Zhao, Yiliu Xu, Yining Liu, Cheng Lu, Yujian Long,
  Yu~Ma, Jinghan Cao, Liang Fan, and Yeyun Xu. 2027.
\newblock \href {https://doi.org/10.1007/978-981-92-7201-3_12} {{GRACE}:
  Grounded reasoning via adapter composition and evidence-aware calibration for
  educational visual question answering}.
\newblock In \emph{Image and Graphics (ICIG 2026)}, volume 17181 of
  \emph{Lecture Notes in Computer Science}, pages 164--175. Springer Nature
  Singapore.

\bibitem[{Li et~al.(2026{\natexlab{c}})Li, Fan, and Zhuang}]{li2026safetyrepro}
Yanhang Li, Zhichao Fan, and Zexin Zhuang. 2026{\natexlab{c}}.
\newblock \href {https://arxiv.org/abs/2605.25492} {{SafetyRepro}:
  Configuration-conditional rank instability on alignment benchmarks}.
\newblock \emph{arXiv preprint arXiv:2605.25492}.

\bibitem[{Liu et~al.(2023)Liu, Wu, Michael, Suhr, West, Koller, Swayamdipta,
  Smith, and Choi}]{liu2023ambient}
Alisa Liu, Zhaofeng Wu, Julian Michael, Alane Suhr, Peter West, Alexander
  Koller, Swabha Swayamdipta, Noah~A. Smith, and Yejin Choi. 2023.
\newblock \href {https://doi.org/10.18653/v1/2023.emnlp-main.51} {We're afraid
  language models aren't modeling ambiguity}.
\newblock In \emph{Proceedings of the 2023 Conference on Empirical Methods in
  Natural Language Processing}, pages 790--807.

\bibitem[{Marmonier et~al.(2025)Marmonier, Bawden, and
  Sagot}]{marmonier2025explicit}
Malik Marmonier, Rachel Bawden, and Beno{\^i}t Sagot. 2025.
\newblock \href {https://doi.org/10.18653/v1/2025.emnlp-main.1599} {Explicit
  learning and the {LLM} in machine translation}.
\newblock In \emph{Proceedings of the 2025 Conference on Empirical Methods in
  Natural Language Processing}, pages 31372--31422.

\bibitem[{Mitchell(1982)}]{mitchell1982generalization}
Tom~M. Mitchell. 1982.
\newblock \href {https://doi.org/10.1016/0004-3702(82)90040-6} {Generalization
  as search}.
\newblock \emph{Artificial Intelligence}, 18(2):203--226.

\bibitem[{Pedregosa et~al.(2011)Pedregosa, Varoquaux, Gramfort, Michel,
  Thirion, Grisel, Blondel, Prettenhofer, Weiss, Dubourg, Vanderplas, Passos,
  Cournapeau, Brucher, Perrot, and Duchesnay}]{pedregosa2011sklearn}
Fabian Pedregosa, Ga{\"e}l Varoquaux, Alexandre Gramfort, Vincent Michel,
  Bertrand Thirion, Olivier Grisel, Mathieu Blondel, Peter Prettenhofer, Ron
  Weiss, Vincent Dubourg, Jake Vanderplas, Alexandre Passos, David Cournapeau,
  Matthieu Brucher, Matthieu Perrot, and {\'E}douard Duchesnay. 2011.
\newblock \href {https://jmlr.org/papers/v12/pedregosa11a.html} {Scikit-learn:
  Machine learning in {P}ython}.
\newblock \emph{Journal of Machine Learning Research}, 12(85):2825--2830.

\bibitem[{{Qwen Team}(2026)}]{qwen2026qwen35}
{Qwen Team}. 2026.
\newblock \href {https://qwen.ai/blog?id=qwen3.5} {{Qwen3.5}: Towards native
  multimodal agents}.
\newblock Qwen blog.

\bibitem[{Ruis et~al.(2020)Ruis, Andreas, Baroni, Bouchacourt, and
  Lake}]{ruis2020gscan}
Laura Ruis, Jacob Andreas, Marco Baroni, Diane Bouchacourt, and Brenden~M.
  Lake. 2020.
\newblock \href
  {https://proceedings.neurips.cc/paper_files/paper/2020/hash/e5a90182cc81e12ab5e72d66e0b46fe3-Abstract.html}
  {A benchmark for systematic generalization in grounded language
  understanding}.
\newblock In \emph{Advances in Neural Information Processing Systems},
  volume~33, pages 19861--19872. Curran Associates, Inc.

\bibitem[{{\c{S}}ahin et~al.(2020){\c{S}}ahin, Kementchedjhieva, Rust, and
  Gurevych}]{sahin2020puzzling}
G{\"o}zde~G{\"u}l {\c{S}}ahin, Yova Kementchedjhieva, Phillip Rust, and Iryna
  Gurevych. 2020.
\newblock \href {https://doi.org/10.18653/v1/2020.acl-main.115} {{P}uzz{L}ing
  machines: A challenge on learning from small data}.
\newblock In \emph{Proceedings of the 58th Annual Meeting of the Association
  for Computational Linguistics}, pages 1241--1254.

\bibitem[{S{\'a}nchez et~al.(2025)S{\'a}nchez, Alastruey, Ropers, Turkatenko,
  Stenetorp, Artetxe, and Costa-juss{\`a}}]{sanchez2024linguini}
Eduardo S{\'a}nchez, Belen Alastruey, Christophe Ropers, Arina Turkatenko,
  Pontus Stenetorp, Mikel Artetxe, and Marta~R. Costa-juss{\`a}. 2025.
\newblock \href {https://doi.org/10.52202/085713-5486} {Linguini: A benchmark
  for language-agnostic linguistic reasoning}.
\newblock In \emph{Advances in Neural Information Processing Systems},
  volume~38, pages 182440--182460. Curran Associates, Inc.
\newblock Datasets and Benchmarks Track.

\bibitem[{Sun et~al.(2026{\natexlab{a}})Sun, Chow, Wang, Liu, Gao, Wu, and
  Kong}]{sun2026scope}
Xian Sun, Wei Chow, Yingshuo Wang, Junhao Liu, Wei Gao, Qing Wu, and Lingdong
  Kong. 2026{\natexlab{a}}.
\newblock \href {https://arxiv.org/abs/2608.06377} {Learning when to trust via
  selective context preference optimization}.
\newblock \emph{arXiv preprint arXiv:2608.06377}.

\bibitem[{Sun et~al.(2026{\natexlab{b}})Sun, Zhou, Xiang, Zhang, Xu, Guan, Ma,
  and Xu}]{sun2026lora}
Yu~Sun, Junwei Zhou, Zuodong Xiang, Yike Zhang, Pengcheng Xu, Xinyu Guan,
  Ruoyun Ma, and Hailu Xu. 2026{\natexlab{b}}.
\newblock \href {https://arxiv.org/abs/2610.12744} {Static gradient attribution
  underperforms a density-matched random mask on loss-based forgetting within
  {LoRA} {B}-matrix}.
\newblock \emph{arXiv preprint arXiv:2610.12744}.

\bibitem[{Tanzer et~al.(2024)Tanzer, Suzgun, Visser, Jurafsky, and
  Melas-Kyriazi}]{tanzer2023mtob}
Garrett Tanzer, Mirac Suzgun, Eline Visser, Dan Jurafsky, and Luke
  Melas-Kyriazi. 2024.
\newblock \href
  {https://proceedings.iclr.cc/paper_files/paper/2024/hash/52d63f9e4b81f866bf69fb3c834aad47-Abstract-Conference.html}
  {A benchmark for learning to translate a new language from one grammar book}.
\newblock In \emph{International Conference on Learning Representations},
  volume 2024, pages 18955--18985.

\bibitem[{Valvoda et~al.(2022)Valvoda, Saphra, Rawski, Williams, and
  Cotterell}]{valvoda2022formal}
Josef Valvoda, Naomi Saphra, Jonathan Rawski, Adina Williams, and Ryan
  Cotterell. 2022.
\newblock \href {https://aclanthology.org/2022.coling-1.525/} {Benchmarking
  compositionality with formal languages}.
\newblock In \emph{Proceedings of the 29th International Conference on
  Computational Linguistics}, pages 6007--6018.

\bibitem[{Wang et~al.(2026{\natexlab{a}})Wang, Ren, Bommireddypalli, Chen,
  Prabhudesai, Zhou, Baral, and Koehn}]{wang2026scitarc}
Hexuan Wang, Yaxuan Ren, Srikar Bommireddypalli, Shuxian Chen, Adarsh
  Prabhudesai, Rongkun Zhou, Elina Baral, and Philipp Koehn.
  2026{\natexlab{a}}.
\newblock \href {https://openreview.net/forum?id=r5iGiq5Ckj} {{SciTaRC}: A
  plan-annotated scientific tabular {QA} benchmark for language reasoning and
  complex computation}.
\newblock In \emph{Third Conference on Language Modeling (COLM)}.

\bibitem[{Wang et~al.(2026{\natexlab{b}})Wang, Zhang, Van~Durme, and
  Khashabi}]{wang2026finer}
Hexuan Wang, Jingyu Zhang, Benjamin Van~Durme, and Daniel Khashabi.
  2026{\natexlab{b}}.
\newblock \href {https://arxiv.org/abs/2604.01432} {Are finer citations always
  better? {R}ethinking granularity for attributed generation}.
\newblock \emph{arXiv preprint arXiv:2604.01432}.

\bibitem[{Wang et~al.(2026{\natexlab{c}})Wang, Wang, Lin, Ji, Wiedemer, Gao,
  Luo, Qian, Huang, Hong, Ge, Ma, He, Zhou, Guo, Mei, Li, Xing, Zhao, Yu, Xiao,
  Jiao, Hou, Zhang, Xu, Zhong, Zhao, Fang, Kitaoka, Xu, Xu, Blacutt, Nguyen,
  Song, Sun, Wen, He, Wang, Wang, Yang, Ma, Milli{\`{e}}re, Shi, Vasconcelos,
  Khashabi, Yuille, Du, Liu, Lin, Liu, Kumar, Li, Yang, Cai, and
  Deng}]{wang2026vbvr}
Maijunxian Wang, Ruisi Wang, Juyi Lin, Ran Ji, Thadd{\"{a}}us Wiedemer,
  Qingying Gao, Dezhi Luo, Yaoyao Qian, Lianyu Huang, Zelong Hong, Jiahui Ge,
  Qianli Ma, Hang He, Yifan Zhou, Lingzi Guo, Lantao Mei, Jiachen Li, Hanwen
  Xing, Tianqi Zhao, and 36 others. 2026{\natexlab{c}}.
\newblock \href {https://proceedings.mlr.press/v306/wang26ia.html} {A very big
  video reasoning suite}.
\newblock In \emph{Proceedings of the 43rd International Conference on Machine
  Learning}, volume 306 of \emph{Proceedings of Machine Learning Research},
  pages 130550--130588. PMLR.

\bibitem[{Wang et~al.(2026{\natexlab{d}})Wang, Wang, Zhang, Wu, Duan, Cai, and
  Zheng}]{wang2026memory}
Yiqi Wang, Yifei Wang, Jiaqi Zhang, Zhangkai Wu, Yiqun Duan, Taotao Cai, and
  Mingkai Zheng. 2026{\natexlab{d}}.
\newblock \href {https://arxiv.org/abs/2610.31446} {When is a memory safe
  enough to act on? {A}ction-conditioned verification for long-lived agents}.
\newblock \emph{arXiv preprint arXiv:2610.31446}.

\bibitem[{Wildenburg et~al.(2024)Wildenburg, Hanna, and
  Pezzelle}]{wildenburg2024dust}
Frank Wildenburg, Michael Hanna, and Sandro Pezzelle. 2024.
\newblock \href {https://doi.org/10.18653/v1/2024.findings-acl.572} {Do
  pre-trained language models detect and understand semantic
  underspecification? {A}sk the {DUST}!}
\newblock In \emph{Findings of the Association for Computational Linguistics:
  ACL 2024}, pages 9598--9613.

\bibitem[{Wolf et~al.(2020)Wolf, Debut, Sanh, Chaumond, Delangue, Moi, Cistac,
  Rault, Louf, Funtowicz, Davison, Shleifer, von Platen, Ma, Jernite, Plu, Xu,
  Le~Scao, Gugger, Drame, Lhoest, and Rush}]{wolf2020transformers}
Thomas Wolf, Lysandre Debut, Victor Sanh, Julien Chaumond, Clement Delangue,
  Anthony Moi, Pierric Cistac, Tim Rault, R{\'e}mi Louf, Morgan Funtowicz, Joe
  Davison, Sam Shleifer, Patrick von Platen, Clara Ma, Yacine Jernite, Julien
  Plu, Canwen Xu, Teven Le~Scao, Sylvain Gugger, and 3 others. 2020.
\newblock \href {https://doi.org/10.18653/v1/2020.emnlp-demos.6} {Transformers:
  State-of-the-art natural language processing}.
\newblock In \emph{Proceedings of the 2020 Conference on Empirical Methods in
  Natural Language Processing: System Demonstrations}, pages 38--45.
  Association for Computational Linguistics.

\bibitem[{Xu et~al.(2024)Xu, Fu, Gao, and Sun}]{xu2024reinvent}
Pengcheng Xu, Tianfan Fu, Wenhao Gao, and Jimeng Sun. 2024.
\newblock \href {https://openreview.net/forum?id=XykiSFid41}
  {{REINVENT}-transformer: Molecular de novo design through transformer-based
  reinforcement learning}.
\newblock In \emph{KDD 2024 Workshop on Artificial Intelligence and Data
  Science for Healthcare (AIDSH)}.

\bibitem[{Xu et~al.(2025)Xu, Shen, Zhou, and Xin}]{xu2025xsz}
Pengcheng Xu, Sicheng Shen, Jieli Zhou, and Hongyi Xin. 2025.
\newblock \href {https://doi.org/10.18653/v1/2025.bionlp-share.33} {Team {XSZ}
  at {B}io{L}ay{S}umm2025: Section-wise summarization, retrieval-augmented
  {LLM}, and reinforcement learning fine-tuning for lay summaries}.
\newblock In \emph{Proceedings of the 24th Workshop on Biomedical Language
  Processing (Shared Tasks)}, pages 275--280, Vienna, Austria. Association for
  Computational Linguistics.

\bibitem[{Xu et~al.(2026)Xu, Li, Lyu, He, Lin, Tang, and Yang}]{xu2026sla}
Yihua Xu, Jize Li, Yan Lyu, Jiani He, You Lin, Shangjing Tang, and Dishu Yang.
  2026.
\newblock \href {https://arxiv.org/abs/2611.26924} {Joint {SLA} admission and
  risk allocation for cloud resource provisioning}.
\newblock \emph{arXiv preprint arXiv:2611.26924}.

\bibitem[{Yang et~al.(2025)Yang, Li, Yang, Zhang, Hui, Zheng, Yu, Gao, Huang,
  Lv, Zheng, Liu, Zhou, Huang, Hu, Ge, Wei, Lin, Tang, Yang, Tu, Zhang, Yang,
  Yang, Zhou, Zhou, Lin, Dang, Bao, Yang, Yu, Deng, Li, Xue, Li, Zhang, Wang,
  Zhu, Men, Gao, Liu, Luo, Li, Tang, Yin, Ren, Wang, Zhang, Ren, Fan, Su,
  Zhang, Zhang, Wan, Liu, Wang, Cui, Zhang, Zhou, and Qiu}]{yang2025qwen3}
An~Yang, Anfeng Li, Baosong Yang, Beichen Zhang, Binyuan Hui, Bo~Zheng, Bowen
  Yu, Chang Gao, Chengen Huang, Chenxu Lv, Chujie Zheng, Dayiheng Liu, Fan
  Zhou, Fei Huang, Feng Hu, Hao Ge, Haoran Wei, Huan Lin, Jialong Tang, and 41
  others. 2025.
\newblock \href {https://doi.org/10.48550/arXiv.2505.09388} {{Qwen3} technical
  report}.
\newblock \emph{arXiv preprint arXiv:2505.09388}.

\bibitem[{Yang et~al.(2024)Yang, Yang, Zhang, Hui, Zheng, Yu, Li, Liu, Huang,
  Wei, Lin, Yang, Tu, Zhang, Yang, Yang, Zhou, Lin, Dang, Lu, Bao, Yang, Yu,
  Li, Xue, Zhang, Zhu, Men, Lin, Li, Tang, Xia, Ren, Ren, Fan, Su, Zhang, Wan,
  Liu, Cui, Zhang, and Qiu}]{yang2025qwen25}
An~Yang, Baosong Yang, Beichen Zhang, Binyuan Hui, Bo~Zheng, Bowen Yu,
  Chengyuan Li, Dayiheng Liu, Fei Huang, Haoran Wei, Huan Lin, Jian Yang,
  Jianhong Tu, Jianwei Zhang, Jianxin Yang, Jiaxi Yang, Jingren Zhou, Junyang
  Lin, Kai Dang, and 23 others. 2024.
\newblock \href {https://doi.org/10.48550/arXiv.2412.15115} {{Qwen2.5}
  technical report}.
\newblock \emph{arXiv preprint arXiv:2412.15115}.

\bibitem[{Zhang et~al.(2024)Zhang, Liu, Lin, and Feng}]{zhang2024teaching}
Chen Zhang, Xiao Liu, Jiuheng Lin, and Yansong Feng. 2024.
\newblock \href {https://doi.org/10.18653/v1/2024.findings-acl.519} {Teaching
  large language models an unseen language on the fly}.
\newblock In \emph{Findings of the Association for Computational Linguistics:
  ACL 2024}, pages 8783--8800.

\bibitem[{Zhang et~al.(2026)Zhang, Xu, Du, Zhang, and Cai}]{zhang2026uopd}
Wenbo Zhang, Pengcheng Xu, Weizhi Du, Jing Zhang, and Hengrui Cai. 2026.
\newblock \href {https://arxiv.org/abs/2609.34036} {{UOPD}: Uncertainty-aware
  intervention for on-policy distillation of multi-turn agents}.
\newblock \emph{arXiv preprint arXiv:2609.34036}.

\bibitem[{Zhang(2026)}]{zhang2026conflict}
Yuchen Zhang. 2026.
\newblock \href {https://arxiv.org/abs/2610.41191} {When {AI} investigators
  cannot trust their tools: Conflict-aware evidence verification for cyber
  threat investigation}.
\newblock \emph{arXiv preprint arXiv:2610.41191}.

\bibitem[{Zhao et~al.(2026)Zhao, Zhao, Guan, Wei, and Deng}]{zhao2026nominate}
Yifan Zhao, Qianyang Zhao, Xinyu Guan, Kai Wei, and Yuming Deng. 2026.
\newblock \href {https://arxiv.org/abs/2610.29391} {Nominate-then-adjudicate:
  {LLM}-assisted one-pass and per-instance {MIP} solver tuning}.
\newblock \emph{arXiv preprint arXiv:2610.29391}.

\bibitem[{Zhao et~al.(2021)Zhao, Wallace, Feng, Klein, and
  Singh}]{zhao2021calibrate}
Zihao Zhao, Eric Wallace, Shi Feng, Dan Klein, and Sameer Singh. 2021.
\newblock \href {https://proceedings.mlr.press/v139/zhao21c.html} {Calibrate
  before use: Improving few-shot performance of language models}.
\newblock In \emph{Proceedings of the 38th International Conference on Machine
  Learning}, volume 139 of \emph{Proceedings of Machine Learning Research},
  pages 12697--12706. PMLR.

\bibitem[{Zheng et~al.(2026)Zheng, Zhai, Cheng, Sha, Huang, Zhou, Qi, Liang,
  Ding, and Xu}]{zheng2026dara}
Tong Zheng, Skylar Zhai, Zhan Cheng, TianMing Sha, Youling Huang, Shuo Zhou,
  Shaotong Qi, Jingcheng Liang, Xuwei Ding, and Pengcheng Xu. 2026.
\newblock \href {https://arxiv.org/abs/2610.00574} {Make sparse rewards count:
  Density-aware reward aggregation for multi-reward {RL}}.
\newblock \emph{arXiv preprint arXiv:2610.00574}.

\end{thebibliography}

\appendix
\section{Formal Semantics and Certificates}
\label{app:proofs}

\paragraph{Proof of evidence-relative necessity.}
Choose $a\neq b$ in $A_t(q,E)$.  By definition, there are
$h_a,h_b\in\version_e(E)$ such that $f_t(h_a,q)=a$ and
$f_t(h_b,q)=b$.  The learner receives the same declared class, evidence, and
query whether the hidden true language is $h_a$ or $h_b$, yet a single payload
cannot be correct in both worlds.  Therefore no learner with only those inputs
can guarantee a correct payload for every compatible true language.
\hfill$\square$

\paragraph{Exact decision algorithm.}
The oracle enumerates all 256 assignments to the ordered lexical--grammar bit
vector, retains the 128 assignments satisfying the public Boolean program,
filters those candidates by the three observed dictionary facts, and executes
the query under each of the 16 survivors.  It returns \texttt{ANSWER} only when
the resulting canonical payload set is a singleton.  Otherwise it returns
\texttt{ABSTAIN} and stores two surviving hypothesis IDs with distinct
payloads.  Hashes bind the canonical program, evidence, query, declared class,
survivor set, answer-support partition, and complete certificate.
Appendix~\ref{app:deduction} and Figure~\ref{fig:deduction} run that algorithm
end to end on the frozen pair of Figure~\ref{fig:main}, so every elimination and
both readouts can be repeated by hand from the prompt alone.

\paragraph{Independent reconstruction.}
The clean-room auditor uses only the Python standard library and imports no
\silica{} package.  It reparses the Boolean DSL, enumerates all programs and
observations, reconstructs the 16 survivors and support partition, and checks
the stored certificates by item ID.  It additionally rebuilds the private
template registry and route schedule, so agreement is not merely a second call
to the generator's oracle.

\section{Four Real Complementary Query Pairs}
\label{app:examples}

Table~\ref{tab:family-examples} reports one immutable same-state pair from each
mechanism family.  Within each row, the hypothesis declaration and projected
observations are identical across the two queries, so Query A and Query B share
exactly the same 16 survivors.  The structured query, response task, and
task-conditioned response schema change together to target the complementary
coordinate.

\begin{table*}[t]
\centering
\scriptsize
\caption{Real \silica{} examples, joined by item ID to evaluator-only gold.
Counts are exact support among the same 16 survivors.  Witness IDs are included
only here for explanation and never enter the model prompt.}
\label{tab:family-examples}
\begin{tabular}{@{}p{2.3cm}p{3.5cm}p{4.2cm}p{4.3cm}@{}}
\toprule
Family & Shared model-visible evidence & Query with unanimous support & Complementary query with disagreement \\
\midrule
Equality chain
& \texttt{nipu$\to$dog}; \texttt{rilu$\to$red};
  \texttt{mabo$\to$see}
& order family $\to$ \texttt{ANSWER(subject\_initial)},
  16/16
& meaning of \texttt{hepu} $\to$ \texttt{ABSTAIN};
  fox 8/16 vs.\ owl 8/16; witnesses h008/h009 \\
\addlinespace
Majority guard
& \texttt{rase$\to$dog}; \texttt{kuhe$\to$see};
  \texttt{toze$\to$red}
& order family $\to$ \texttt{ANSWER(subject\_initial)},
  16/16
& meaning of \texttt{puso} $\to$ \texttt{ABSTAIN};
  fox 8/16 vs.\ owl 8/16; witnesses h014/h015 \\
\addlinespace
Parity guard
& \texttt{pudo$\to$red}; \texttt{levu$\to$see};
  \texttt{reze$\to$dog}
& meaning of \texttt{helu} $\to$ \texttt{ANSWER(fox)},
  16/16
& order family $\to$ \texttt{ABSTAIN};
  subject-initial 8/16 vs.\ verb-initial 8/16;
  witnesses h002/h018 \\
\addlinespace
Selector multiplexer
& \texttt{devi$\to$dog}; \texttt{babe$\to$see};
  \texttt{naga$\to$red}
& meaning of \texttt{mivu} $\to$ \texttt{ANSWER(fox)},
  16/16
& order family $\to$ \texttt{ABSTAIN};
  subject-initial 8/16 vs.\ verb-initial 8/16;
  witnesses h004/h020 \\
\bottomrule
\end{tabular}
\end{table*}

For the parity pair visualized in Figure~\ref{fig:main}, the two exact response
objects are
\begin{center}
\small\ttfamily
\{"decision":"ANSWER","answer":"fox"\}\\
\{"decision":"ABSTAIN","answer":null\}.
\end{center}
Any prose, Markdown fence, additional key, wrong payload type, or non-JSON
value is invalid.  For the underdetermined row, the hidden sampled target happens
to be subject-initial, but that field is not an answer: the other eight
compatible languages are verb-initial.  This distinction is why target
accuracy is not the benchmark objective.

\subsection{Complete renderer messages for Figure~\ref{fig:main}}
\label{app:complete-prompts}

Figure~\ref{fig:main} uses a readable, faithful excerpt.  The files below are
generated directly from the frozen renderer.  Machine-exact JSON arrays are
checked in beside these losslessly wrapped views.  To reconstruct each exact
message string, concatenate the content lines between its
\texttt{BEGIN}/\texttt{END} markers and remove only the display line feeds;
the wrapper also records a content SHA-256.  Both rows belong to
\texttt{hard-v2-state-}\allowbreak\texttt{b1e9ee8eb2baed19aee4}.

\paragraph{Complete ANSWER input.}
Item \nolinkurl{silica-d8de6e0dbb2c8bd57d7c8bd9} has prompt SHA-256
\texttt{7d29fc3adf69506064c959c3996a661d}\allowbreak
\texttt{2d881e1f67b83abee63e3f77682a61cb}.

\begin{Verbatim}
LOSSLESS DISPLAY: concatenate content lines between BEGIN/END; remove only LF.
The BEGIN/END wrapper lines are metadata and are not part of message content.
BEGIN MESSAGE 0 role=system content_sha256=b10f321447f2fbe3b973a8fc079908a2595ea5aefd23028bded201395d9a6e27
You are solving the SILICA-Hard posterior-matched paired-query identifiability reasoning
 benchmark. The supplied declaration defines an exact finite hypothesis class: enumerate
 all Boolean feature assignments, retain only assignments accepted by every declared Boo
lean-program constraint, and then retain only assignments consistent with every observed
 form/meaning pair. Evaluate the query under every surviving assignment. Return ANSWER o
nly if all survivors produce one identical canonical answer; otherwise return ABSTAIN. R
eason silently. Emit exactly one JSON object satisfying canonical_response_schema, with 
no markdown, rationale, confidence, or additional keys.
END MESSAGE 0
BEGIN MESSAGE 1 role=user content_sha256=9136ef53746dacd5212d3d3db6b8f8850ce946b70a014f4a4c69e284ac8909ee
{"adapter_protocol_hash":"eb211b7350e62a6eadae8f3f5ce73d7e1536cc10c9984e342763989394f84c
5f","adapter_task":"hard_posterior_matched","adapter_version":"silica-hard-pair-aware-ad
apter-v2","canonical_response_schema":{"oneOf":[{"additionalProperties":false,"propertie
s":{"answer":{"enum":["fox","owl","dog","bear"],"type":"string"},"decision":{"enum":["AN
SWER"],"type":"string"}},"required":["decision","answer"],"type":"object"},{"additionalP
roperties":false,"properties":{"answer":{"type":"null"},"decision":{"enum":["ABSTAIN"],"
type":"string"}},"required":["decision","answer"],"type":"object"}],"type":"object"},"de
cision_rule":{"ABSTAIN":"at least two observation-compatible hypotheses have different c
anonical query answers","ANSWER":"every observation-compatible hypothesis has the same c
anonical query answer"},"hypothesis_declaration":{"base_class_size":256,"boolean_dsl_sem
antics":{"canonicalization":{"arities":{"and":"at least 2","eq":2,"implies":2,"ite":3,"n
ot":1,"or":"at least 2","xor":"at least 2"},"commutative_argument_order":"lexicographic 
canonical JSON ordering for and/or/xor/eq","constraint_order":"lexicographic canonical J
SON ordering","object_keys":"lexicographic canonical JSON ordering"},"domain":"Boolean v
alues false/true, encoded by SILICA feature bits 0/1","operators":{"and":"true iff every
 arg is true","const":"the literal Boolean value","eq":"true iff its two args have the s
ame Boolean value","implies":"equivalent to (not first) or second","ite":"second arg if 
first arg is true, otherwise third arg","not":"Boolean negation of arg","or":"true iff a
t least one arg is true","var":"the Boolean bit of the named feature","xor":"true iff an
 odd number of args are true"},"program_acceptance":"a hypothesis is admitted iff every 
constraint evaluates to true"},"boolean_program":{"constraints":[{"args":[{"args":[{"arg
":{"args":[{"args":[{"name":"modifier_swap_01","op":"var"},{"name":"noun_swap_23","op":"
var"},{"name":"verb_swap_01","op":"var"}],"op":"xor"},{"args":[{"name":"modifier_swap_01
","op":"var"},{"name":"verb_swap_01","op":"var"}],"op":"eq"}],"op":"and"},"op":"not"},{"
args":[{"args":[{"name":"verb_swap_01","op":"var"},{"name":"noun_swap_23","op":"var"},{"
arg":{"name":"modifier_swap_01","op":"var"},"op":"not"}],"op":"ite"},{"name":"order_fami
ly","op":"var"}],"op":"eq"}],"op":"implies"},{"args":[{"args":[{"args":[{"name":"modifie
r_swap_01","op":"var"},{"name":"noun_swap_23","op":"var"},{"name":"verb_swap_01","op":"v
ar"}],"op":"xor"},{"args":[{"name":"modifier_swap_01","op":"var"},{"name":"verb_swap_01"
,"op":"var"}],"op":"eq"}],"op":"and"},{"args":[{"args":[{"name":"noun_swap_23","op":"var
"},{"name":"verb_swap_01","op":"var"},{"name":"modifier_swap_01","op":"var"}],"op":"ite"
},{"name":"noun_swap_01","op":"var"}],"op":"eq"}],"op":"implies"}],"op":"and"}],"dsl_ver
sion":"silica-boolean-constraint-program-v1","variables":["noun_swap_01","noun_swap_23",
"verb_swap_01","modifier_swap_01","order_family","order_variant","alignment","agreement"
]},"class_size":128,"constant_markers":{"agreement_plural":"poso","agreement_singular":"
sena","case_slot_a":"hopo","case_slot_b":"tabu","past_suffix":"vagi","plural_suffix":"pe
pu","preverbal_negation":"dare"},"constant_rules":["modifiers precede nouns","plural and
 case are suffixes","past precedes agreement on the verb"],"exact_enumeration":{"allowli
st_required":false,"filter":"retain assignments satisfying every Boolean-program constra
int","universe":"all 2^8 assignments in feature_variables_lsb_to_msb order"},"feature_va
riables_lsb_to_msb":[{"kind":"lexical","name":"noun_swap_01","values":["identity","swapp
ed"]},{"kind":"lexical","name":"noun_swap_23","values":["identity","swapped"]},{"kind":"
lexical","name":"verb_swap_01","values":["identity","swapped"]},{"kind":"lexical","name"
:"modifier_swap_01","values":["identity","swapped"]},{"kind":"grammar","name":"order_fam
ily","values":["subject_initial","verb_initial"]},{"kind":"grammar","name":"order_varian
t","values":["variant_a","variant_b"]},{"kind":"grammar","name":"alignment","values":["n
ominative_accusative","ergative_absolutive"]},{"kind":"grammar","name":"agreement","valu
es":["zero","agent_number"]}],"fixed_lexicon":[{"category":"verb","form":"petu","meaning
":"chase"},{"category":"verb","form":"mizu","meaning":"sleep","valency":"intransitive"},
{"category":"modifier","form":"zabu","meaning":"green"},{"category":"modifier","form":"n
upo","meaning":"small"}],"grammar_axes":[{"name":"order_family","values":["subject_initi
al","verb_initial"]},{"name":"order_variant","values":["variant_a","variant_b"]},{"name"
:"alignment","values":["nominative_accusative","ergative_absolutive"]},{"name":"agreemen
t","values":["zero","agent_number"]}],"grammar_semantics":{"agreement_mapping":{"agent_n
umber":{"plural":"agreement_plural","singular":"agreement_singular"},"zero":"no_agreemen
t_suffix"},"alignment_mapping":{"ergative_absolutive":{"intransitive_subject_S":"case_sl
ot_b","transitive_agent_A":"case_slot_a","transitive_patient_O":"case_slot_b"},"nominati
ve_accusative":{"intransitive_subject_S":"case_slot_a","transitive_agent_A":"case_slot_a
","transitive_patient_O":"case_slot_b"}},"intransitive_order":{"subject_initial":"SV","v
erb_initial":"VS"},"order_mapping":[{"order_family":"subject_initial","order_variant":"v
ariant_a","word_order":"SVO"},{"order_family":"subject_initial","order_variant":"variant
_b","word_order":"SOV"},{"order_family":"verb_initial","order_variant":"variant_a","word
_order":"VSO"},{"order_family":"verb_initial","order_variant":"variant_b","word_order":"
VOS"}]},"lexical_axes":[{"category":"noun","constraint":"either_bijection","forms":["hel
u","vuga"],"name":"noun_pair_01","possible_meanings":["fox","owl"]},{"category":"noun","
constraint":"either_bijection","forms":["kuri","reze"],"name":"noun_pair_23","possible_m
eanings":["dog","bear"]},{"category":"verb","constraint":"either_bijection","forms":["le
vu","bati"],"name":"verb_pair_01","possible_meanings":["see","help"]},{"category":"modif
ier","constraint":"either_bijection","forms":["pudo","gisu"],"name":"modifier_pair_01","
possible_meanings":["red","blue"]}],"schema_family":"silica-hard-posterior-matched-v2","
type":"closed_world_posterior_matched_boolean_program_subset"},"observations":[{"categor
y":"modifier","channel":"dictionary","form":"pudo","semantic":"red"},{"category":"verb",
"channel":"dictionary","form":"levu","semantic":"see"},{"category":"noun","channel":"dic
tionary","form":"reze","semantic":"dog"}],"prompt_version":"silica-hard-posterior-matche
d-json-v2","query":{"category":"noun","form":"helu"},"response_schema_version":"silica-a
nswer-json-schema-v2","response_task":"lexicon"}
END MESSAGE 1
\end{Verbatim}

\paragraph{Exact ANSWER gold object.}
\begin{Verbatim}
{
  "answer": "fox",
  "decision": "ANSWER"
}
\end{Verbatim}

\paragraph{Complete ABSTAIN input.}
Item \nolinkurl{silica-4df031b78b218b09cff975e2} has prompt SHA-256
\texttt{5a1cb07dc66456f83b275de957a04386}\allowbreak
\texttt{26ec024da9e5c14d74209ca94b2ce414}.

\begin{Verbatim}
LOSSLESS DISPLAY: concatenate content lines between BEGIN/END; remove only LF.
The BEGIN/END wrapper lines are metadata and are not part of message content.
BEGIN MESSAGE 0 role=system content_sha256=b10f321447f2fbe3b973a8fc079908a2595ea5aefd23028bded201395d9a6e27
You are solving the SILICA-Hard posterior-matched paired-query identifiability reasoning
 benchmark. The supplied declaration defines an exact finite hypothesis class: enumerate
 all Boolean feature assignments, retain only assignments accepted by every declared Boo
lean-program constraint, and then retain only assignments consistent with every observed
 form/meaning pair. Evaluate the query under every surviving assignment. Return ANSWER o
nly if all survivors produce one identical canonical answer; otherwise return ABSTAIN. R
eason silently. Emit exactly one JSON object satisfying canonical_response_schema, with 
no markdown, rationale, confidence, or additional keys.
END MESSAGE 0
BEGIN MESSAGE 1 role=user content_sha256=3b456bde71620707fde7be24f4db44f7e9be460f504c279dc1383a3c8f43cd69
{"adapter_protocol_hash":"eb211b7350e62a6eadae8f3f5ce73d7e1536cc10c9984e342763989394f84c
5f","adapter_task":"hard_posterior_matched","adapter_version":"silica-hard-pair-aware-ad
apter-v2","canonical_response_schema":{"oneOf":[{"additionalProperties":false,"propertie
s":{"answer":{"enum":["subject_initial","verb_initial"],"type":"string"},"decision":{"en
um":["ANSWER"],"type":"string"}},"required":["decision","answer"],"type":"object"},{"add
itionalProperties":false,"properties":{"answer":{"type":"null"},"decision":{"enum":["ABS
TAIN"],"type":"string"}},"required":["decision","answer"],"type":"object"}],"type":"obje
ct"},"decision_rule":{"ABSTAIN":"at least two observation-compatible hypotheses have dif
ferent canonical query answers","ANSWER":"every observation-compatible hypothesis has th
e same canonical query answer"},"hypothesis_declaration":{"base_class_size":256,"boolean
_dsl_semantics":{"canonicalization":{"arities":{"and":"at least 2","eq":2,"implies":2,"i
te":3,"not":1,"or":"at least 2","xor":"at least 2"},"commutative_argument_order":"lexico
graphic canonical JSON ordering for and/or/xor/eq","constraint_order":"lexicographic can
onical JSON ordering","object_keys":"lexicographic canonical JSON ordering"},"domain":"B
oolean values false/true, encoded by SILICA feature bits 0/1","operators":{"and":"true i
ff every arg is true","const":"the literal Boolean value","eq":"true iff its two args ha
ve the same Boolean value","implies":"equivalent to (not first) or second","ite":"second
 arg if first arg is true, otherwise third arg","not":"Boolean negation of arg","or":"tr
ue iff at least one arg is true","var":"the Boolean bit of the named feature","xor":"tru
e iff an odd number of args are true"},"program_acceptance":"a hypothesis is admitted if
f every constraint evaluates to true"},"boolean_program":{"constraints":[{"args":[{"args
":[{"arg":{"args":[{"args":[{"name":"modifier_swap_01","op":"var"},{"name":"noun_swap_23
","op":"var"},{"name":"verb_swap_01","op":"var"}],"op":"xor"},{"args":[{"name":"modifier
_swap_01","op":"var"},{"name":"verb_swap_01","op":"var"}],"op":"eq"}],"op":"and"},"op":"
not"},{"args":[{"args":[{"name":"verb_swap_01","op":"var"},{"name":"noun_swap_23","op":"
var"},{"arg":{"name":"modifier_swap_01","op":"var"},"op":"not"}],"op":"ite"},{"name":"or
der_family","op":"var"}],"op":"eq"}],"op":"implies"},{"args":[{"args":[{"args":[{"name":
"modifier_swap_01","op":"var"},{"name":"noun_swap_23","op":"var"},{"name":"verb_swap_01"
,"op":"var"}],"op":"xor"},{"args":[{"name":"modifier_swap_01","op":"var"},{"name":"verb_
swap_01","op":"var"}],"op":"eq"}],"op":"and"},{"args":[{"args":[{"name":"noun_swap_23","
op":"var"},{"name":"verb_swap_01","op":"var"},{"name":"modifier_swap_01","op":"var"}],"o
p":"ite"},{"name":"noun_swap_01","op":"var"}],"op":"eq"}],"op":"implies"}],"op":"and"}],
"dsl_version":"silica-boolean-constraint-program-v1","variables":["noun_swap_01","noun_s
wap_23","verb_swap_01","modifier_swap_01","order_family","order_variant","alignment","ag
reement"]},"class_size":128,"constant_markers":{"agreement_plural":"poso","agreement_sin
gular":"sena","case_slot_a":"hopo","case_slot_b":"tabu","past_suffix":"vagi","plural_suf
fix":"pepu","preverbal_negation":"dare"},"constant_rules":["modifiers precede nouns","pl
ural and case are suffixes","past precedes agreement on the verb"],"exact_enumeration":{
"allowlist_required":false,"filter":"retain assignments satisfying every Boolean-program
 constraint","universe":"all 2^8 assignments in feature_variables_lsb_to_msb order"},"fe
ature_variables_lsb_to_msb":[{"kind":"lexical","name":"noun_swap_01","values":["identity
","swapped"]},{"kind":"lexical","name":"noun_swap_23","values":["identity","swapped"]},{
"kind":"lexical","name":"verb_swap_01","values":["identity","swapped"]},{"kind":"lexical
","name":"modifier_swap_01","values":["identity","swapped"]},{"kind":"grammar","name":"o
rder_family","values":["subject_initial","verb_initial"]},{"kind":"grammar","name":"orde
r_variant","values":["variant_a","variant_b"]},{"kind":"grammar","name":"alignment","val
ues":["nominative_accusative","ergative_absolutive"]},{"kind":"grammar","name":"agreemen
t","values":["zero","agent_number"]}],"fixed_lexicon":[{"category":"verb","form":"petu",
"meaning":"chase"},{"category":"verb","form":"mizu","meaning":"sleep","valency":"intrans
itive"},{"category":"modifier","form":"zabu","meaning":"green"},{"category":"modifier","
form":"nupo","meaning":"small"}],"grammar_axes":[{"name":"order_family","values":["subje
ct_initial","verb_initial"]},{"name":"order_variant","values":["variant_a","variant_b"]}
,{"name":"alignment","values":["nominative_accusative","ergative_absolutive"]},{"name":"
agreement","values":["zero","agent_number"]}],"grammar_semantics":{"agreement_mapping":{
"agent_number":{"plural":"agreement_plural","singular":"agreement_singular"},"zero":"no_
agreement_suffix"},"alignment_mapping":{"ergative_absolutive":{"intransitive_subject_S":
"case_slot_b","transitive_agent_A":"case_slot_a","transitive_patient_O":"case_slot_b"},"
nominative_accusative":{"intransitive_subject_S":"case_slot_a","transitive_agent_A":"cas
e_slot_a","transitive_patient_O":"case_slot_b"}},"intransitive_order":{"subject_initial"
:"SV","verb_initial":"VS"},"order_mapping":[{"order_family":"subject_initial","order_var
iant":"variant_a","word_order":"SVO"},{"order_family":"subject_initial","order_variant":
"variant_b","word_order":"SOV"},{"order_family":"verb_initial","order_variant":"variant_
a","word_order":"VSO"},{"order_family":"verb_initial","order_variant":"variant_b","word_
order":"VOS"}]},"lexical_axes":[{"category":"noun","constraint":"either_bijection","form
s":["helu","vuga"],"name":"noun_pair_01","possible_meanings":["fox","owl"]},{"category":
"noun","constraint":"either_bijection","forms":["kuri","reze"],"name":"noun_pair_23","po
ssible_meanings":["dog","bear"]},{"category":"verb","constraint":"either_bijection","for
ms":["levu","bati"],"name":"verb_pair_01","possible_meanings":["see","help"]},{"category
":"modifier","constraint":"either_bijection","forms":["pudo","gisu"],"name":"modifier_pa
ir_01","possible_meanings":["red","blue"]}],"schema_family":"silica-hard-posterior-match
ed-v2","type":"closed_world_posterior_matched_boolean_program_subset"},"observations":[{
"category":"modifier","channel":"dictionary","form":"pudo","semantic":"red"},{"category"
:"verb","channel":"dictionary","form":"levu","semantic":"see"},{"category":"noun","chann
el":"dictionary","form":"reze","semantic":"dog"}],"prompt_version":"silica-hard-posterio
r-matched-json-v2","query":{"feature":"order_family"},"response_schema_version":"silica-
answer-json-schema-v2","response_task":"grammar_feature"}
END MESSAGE 1
\end{Verbatim}

\paragraph{Exact ABSTAIN gold object.}
\begin{Verbatim}
{
  "answer": null,
  "decision": "ABSTAIN"
}
\end{Verbatim}

The ANSWER prompt's task, structured query, and schema request a noun meaning;
the ABSTAIN prompt's corresponding fields request a grammar feature.  Their
hypothesis declaration and projected observations are identical.  Neither
prompt contains survivor counts, support partitions, witnesses, or gold.

\subsection{Field boundary}

\begin{table}[t]
\centering
\scriptsize
\caption{The renderer enforces a strict public/private boundary.}
\label{tab:visibility}
\begin{tabular}{@{}p{0.46\columnwidth}p{0.46\columnwidth}@{}}
\toprule
Model-visible & Evaluator-only \\
\midrule
Boolean DSL semantics and program; feature values and realization rules;
three dictionary facts; query; decision rule; exact response schema
&
Family, route, template and pair metadata; observation IDs; semantic and
structural hashes; hidden role orientation; target language and target answer;
gold decision and payload; survivors, supports, witnesses, and certificates \\
\bottomrule
\end{tabular}
\end{table}

Public files contain routing fields needed to assemble complete cohorts.  The
prompt renderer explicitly projects them away; the evaluator joins private
gold only after the public cohort and its hash are frozen.

\section{Guard Families and Semantic Packing}
\label{app:registry}

\begin{figure*}[t]
  \centering
  \includegraphics[width=\textwidth]{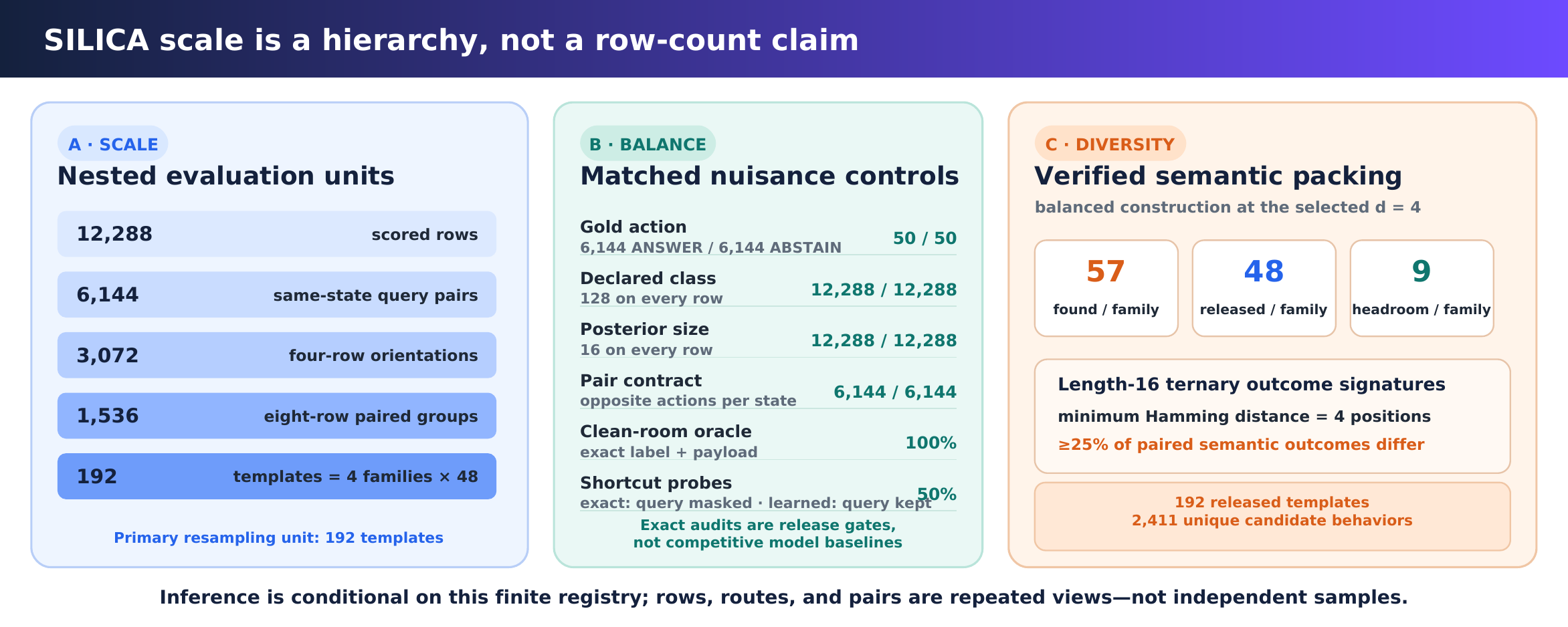}
  \caption{\textbf{Rows are nested views, not independent examples.}
  Panel A reports the exact evaluation hierarchy; inference resamples
  192 semantic templates within four 48-template mechanism families.  Panel B
  shows exact nuisance balance and shortcut gates.  Panel C reports the
  sealed packing at the selected $d=4$: the search found 57 templates per
  family, released 48, and retained nine of measured headroom.  The
  length-16 ternary signatures differ in at least four positions.  These are
  properties of one fixed finite registry, not a population estimate or an
  unreported search frontier.}
  \label{fig:dataset-geometry}
\end{figure*}

Let $g(X)$ be a guard over the three observed bits, let $Q$ and $N$ be the two
query-facing coordinates, and let $p(X)$ and $n(X)$ be their payload
expressions.  Table~\ref{tab:family-skeletons} gives the exact placeholder
skeletons before variable permutation and canonicalization.  All four encode
the same switch---$g$ constrains $Q$, while $\neg g$ constrains $N$---through
different Boolean AST structures.

\begin{table*}[t]
\centering
\scriptsize
\caption{Four structural encodings of the guarded-switch principle.}
\label{tab:family-skeletons}
\begin{tabular}{@{}lp{9.2cm}p{2.5cm}@{}}
\toprule
Family & Canonical constraint skeleton & Characteristic root \\
\midrule
Parity guard
& $(g\Rightarrow Q=p)\land(\neg g\Rightarrow N=n)$
& implication \\
Majority guard
& $(\neg g\lor Q=p)\land(g\lor N=n)$
& disjunction \\
Equality chain
& $[(g\land Q)=(g\land p)]\land[(\neg g\land N)=(\neg g\land n)]$
& equality \\
Selector multiplexer
& $\operatorname{ITE}(g,Q=p,N=n)$
& if--then--else \\
\bottomrule
\end{tabular}
\end{table*}

Candidate enumeration produces 914 parity, 1,112 majority, 268 equality, and
350 selector programs.  Their union contains 2,411 globally unique paired
behaviors.  Filtering against the fixed 12-template support registry at
length-16 ternary paired-signature distance at least four positions leaves
854, 1,021, 229, and 340 candidates, respectively.  A standard-library
deterministic search checks all
24 family orders under two fixed conflict-degree scores.  The best complete
balanced cycle contains 57 templates per family; the release selects 48.

The paired signature records the outcome of both query roles in each of the
eight observed-bit states as \texttt{ABSTAIN}, \texttt{ANSWER:0}, or
\texttt{ANSWER:1}.  Evaluation--evaluation and evaluation--support minimum
distance is four under both this signature and the derived full-class truth
table.  All 192 evaluation template semantics are unique.

\subsection{Routes and complete units}

For each template, route $j$ selects sorted member
$j\bmod|\text{stratum}|$ from each of four strata: active/inactive guard crossed
with payload 0/1.  A route contains four states and both query roles, forming
one eight-row paired group.  Across eight routes, every template covers all
eight observed-bit states; each stratum contributes exactly eight selections,
and members receive their floor-or-ceiling share.  The common model cohort
contains one route per template.  Only the full cohort contains all eight
routes per template.

\section{Construction and Shortcut Audits}
\label{app:audits}

\begin{table}[t]
\centering
\scriptsize
\caption{Frozen release checks, all over the complete 12,288-row artifact.}
\label{tab:appendix-audits}
\begin{tabular}{@{}lr@{}}
\toprule
Check & Exact result \\
\midrule
In-generator label/payload oracle & 12,288/12,288 \\
Independent clean-room oracle & 12,288/12,288 \\
Declared class size 128 & 12,288/12,288 \\
Posterior size 16 & 12,288/12,288 \\
Same-state opposite-action pairs & 6,144/6,144 \\
Unique rendered prompt hashes & 12,288/12,288 \\
Reverse-order ID join & identical \\
Query-masked raw-state control & 6,144/12,288 \\
Word, held-out-template & 6,144/12,288 \\
Character, held-out-template & 6,144/12,288 \\
Word, held-out-family & 6,144/12,288 \\
Character, held-out-family & 6,144/12,288 \\
\bottomrule
\end{tabular}
\end{table}

The learned probes remove identifiers, hashes, routes, and observation surface
forms while retaining the public query and Boolean program.  Template folds
are grouped eight ways and family folds four ways; each of the four
predeclared probes is exactly at the 0.500 acceptance gate.  These audits
exclude the tested shortcut families, not every possible heuristic.

\section{Metrics, Uncertainty, and Training Signals}
\label{app:metrics}

Every report includes exact numerators and denominators for strict joint
accuracy, balanced strict joint accuracy, decision accuracy, balanced decision
accuracy, ANSWER and ABSTAIN recall, exact payload recall, answer coverage,
format validity, false certainty, false abstention, and request errors.
Breakdowns retain mechanism family, public query axis, and route.

At the four-row level, the evaluator requires the complete oriented decision
pattern and, separately, both answerable payloads.  At the paired level, it
requires both orientations, all eight strict responses, and all four
same-state query-complement edges.  Unoriented complement and factorial
consistency are structural diagnostics only: a globally inverted predictor can
satisfy them.

Intervals use 10,000 percentile replicates with seed 20260803.  Each replicate
resamples the 48 semantic templates within each of the four mechanism families
and moves every associated route, state, and query together.  The intervals
therefore describe sensitivity over this finite registry; they are neither
independent-row confidence intervals nor hypothesis tests.

\paragraph{Reference levels for the two derived diagnostics.}
The decision signal $\mathrm{AR}-\mathrm{FCR}$ and conditional payload accuracy
$\mathrm{CPA}=\mathrm{EAR}/\mathrm{AR}$ are reported because AR, UR and EAR each
mix a model's chosen answer coverage with its actual competence.  Consider a
predictor that emits \texttt{ANSWER} with probability $\alpha$ independently of
the row.  It attains $\mathrm{AR}=\mathrm{FCR}=\alpha$, hence decision signal
$0$ and, on format-valid rows, $\mathrm{BDA}=1/2$ for every $\alpha$; its
$\mathrm{UR}=1-\alpha$ and $\mathrm{EAR}=\alpha/2$ nevertheless sweep the entire
range as $\alpha$ varies, and its BJA sweeps $[.25,.50]$.  Reporting UR, EAR or
BJA alone therefore cannot distinguish a competent model from a differently
tuned constant.  The CPA reference of exactly $1/2$ is a property of the
construction rather than an asymptotic argument: in all 3{,}072 four-cell units
the two answerable rows carry different canonical payloads, verified directly
against the sealed gold records.

\paragraph{Sealed model ledger.}
Model numbers in the main text are not transcribed.  A collector reverifies each
archive (zero-error completion status, UUID agreement, dataset release identity,
cohort identity, per-file SHA-256 against the completion record, evaluator report
hash, bootstrap replicate count and seed, declared row count, mechanism-family
strata, and runtime node) and emits one canonical ledger carrying its own
payload digest.  The panel table and panel figure are both generated from that
ledger, and the figure script recomputes the digest and refuses to draw on
mismatch.  Any archive that is partial, nonzero-error, or hash-inconsistent is
rejected rather than downweighted.

\paragraph{RL-ready, not an RL result.}
The official scorer yields deterministic row rewards for schema validity,
decision correctness, and exact joint correctness, plus sparse all-four and
all-eight group rewards; combining them is a multi-reward problem in which
sparse components can be under-weighted \citep{zheng2026dara}.  These
components can support supervised learning, rejection sampling, or
reinforcement learning without a judge model, much as scored rewards drive RL
fine-tuning for molecular design and lay summarization
\citep{xu2024reinvent,xu2025xsz}.  Private support partitions and witnesses can
additionally supervise a future certificate-producing agent, for example as
teacher guidance in on-policy distillation \citep{zhang2026uopd,huang2026guide}.  The current direct response contract requests
only decision and payload, and this paper reports no policy optimization,
training curve, agent trajectory, or transfer claim.

\section{Direct-Model and Provenance Protocol}
\label{app:protocol}

The frozen common panel evaluates Qwen2.5-Instruct at 0.5B, 1.5B, 3B, 7B, 14B,
and 32B parameters, Phi-3.5-mini-instruct, and SmolLM2-1.7B-Instruct on the same
1,536 rows.  Qwen2.5-14B additionally runs on all 12,288 rows; Qwen2.5-32B is
the stronger common-cohort sentinel; and Qwen2.5-1.5B receives one ID-joined
cross-hardware replicate.  Because open-weight checkpoints are often
fine-tuned, merged and re-released \citep{li2026witness}, revisions are
immutable 40-character repository commits.  An additional reasoning model may
be reported only under a separately frozen, complete, provenance-valid
condition; exploratory or partial output is never merged with the panel.

Each condition in the frozen original local panel uses one CUDA GPU, bfloat16,
batch size two, greedy decoding, temperature and seed zero, at most 384 new
tokens, left padding, the tokenizer chat template, and local files only.
Tensor-parallel extensions are separate frozen conditions.  Output must terminate with EOS.
The parser preserves generated special-token IDs and rejects any special token
inside the content span, a missing EOS, cap termination, prose, repair, or
additional JSON keys.

Each local archive binds the dataset and public cohort, post-selection gold
audit, prompt and schema, model and tokenizer revisions, a safetensors-only
load plan, every weight/config/tokenizer file hash and size, effective
generation configuration, source commit and executed source hashes, isolated
runtime, hardware, token IDs, request hashes, and a canonical run UUID.  Weight
index paths must be normalized relative POSIX safetensors paths contained in
the sealed manifest.  Snapshot files remain descriptor-sealed through loading
and are revalidated before output is appended.  Any model, source, cache, or
configuration drift aborts rather than silently defining a new condition.

\paragraph{Trace-permitted extension conditions.}
The trace-permitted conditions of Section~\ref{sec:results} are served by vLLM
0.18.1 \citep{kwon2023vllm} through the tokenizer chat template; thinking is
toggled by the template's \texttt{enable\_thinking} switch, the
\texttt{deepseek\_r1} reasoning parser separates the trace from the final
content, and only the final content is parsed and scored.  Within each matched
pair the checkpoint files, revision, prompt messages, sampling settings,
16{,}384-token budget, seed and engine version are identical.  The Qwen3-1.7B
and Qwen3-235B pairs differ only in the thinking switch and the reasoning
parser; both 235B servers load the same GPTQ checkpoint with the same Marlin
GPTQ kernel, as their engine logs record.  Four-bit quantization can lower task
performance \citep{haque2025quantization}, so the 235B results describe this
Int4 checkpoint, not the full-precision model.  The Qwen3.5-2B decoding pair differs
only in temperature and top-$k$.  The two arms of each pair ran from different
revisions of our evaluation harness; the changes concern how a response without
final content is recorded (as an incorrect row rather than a request error) and
how a run is admitted to its frozen configuration, not the request sent to the
model.  The Qwen3.5-0.8B and Qwen3.5-2B trace-on arms each lose one row to an
adapter error in our harness, so the evaluator does not finalize them; we report
only their termination rates and their strict format validity, computed with
the evaluator's parser with the lost row counted as invalid.

\paragraph{Phi-3.5 with a multi-token stop contract.}
Under the local panel's single-EOS contract every Phi-3.5-mini response carries
its end-of-turn token inside the scored span (1{,}536/1{,}536), so the sealed
condition has no valid rows.  A replacement condition keeps the checkpoint,
revision, cohort, prompt, response schema and scoring, and serves the model
through vLLM, which stops on the checkpoint's declared end-of-turn tokens;
decoding is greedy with a 16{,}384-token budget.  It reached 314 of the 1{,}536
rows within its time limit.  All 126 responses that stop do so on the
end-of-turn token, so the original artifact cannot recur, but only 49 of the 314
rows are strictly format-valid: 188 rows repeat one line until the token cap,
and 77 of the 126 stopped rows break the output contract (35 surround the JSON
object with prose, 13 wrap it in a code fence, and 29 are bare objects the
strict parser rejects).  A complete run could therefore not exceed
$(49+1{,}222)/1{,}536<.83$, below the .95 gate, and the condition remains
unevaluable.

Hosted inference is a separate stratum.  Provider routing is pinned,
fallbacks are disabled, the credential-free canonical request body is hashed,
and request failures remain scored errors.  A hosted result can enter a table
only as a complete 1,536-row archive; quota-limited canaries do not receive a
paper score.

\section{Reproducibility and Data Statement}
\label{app:reproducibility}

\paragraph{Release.}
The code (generator, validators, clean-room oracle, prompt-only solver, shortcut
audits and evaluator) is released under the Apache-2.0 license at \codeurl.
The dataset, with every model response analyzed in this paper, is released under
CC BY 4.0 at \dataurl; the model responses remain subject to the license of the
model that produced them.  The released files are byte-identical to the
evaluated artifact.  A few identifiers inside them predate the final name and
are kept so that every hash below and every released response still matches:
the version field reads \texttt{0.2.0-rc1}, grouping identifiers carry the
prefix \texttt{hard-v2-}, and the model-facing system message and adapter
string name the benchmark ``SILICA-Hard''.

\paragraph{Identities.}
The released artifact has the following SHA-256 identities.

\noindent{\scriptsize
\begin{tabular}{@{}ll@{}}
Dataset aggregate & \texttt{d956532e42ecc666835f92071d779c7a} \\
 & \texttt{d646b707b859794ffd91175d8d954b57} \\
Strict release & \texttt{f49ab76b635425db3ce60d69e3c2f13a} \\
 & \texttt{204f98a3f1e278778f7008570955c669} \\
Root manifest & \texttt{22032191c922fbbc424bfce0a7fe3dd9} \\
 & \texttt{96d040764dbc472c0e7432b5b4b58a15} \\
Public manifest & \texttt{01f0ae747baff9e04c6ef648bf461848} \\
 & \texttt{aded6636921f8d49a5b291db864d1c02} \\
Construction & \texttt{5f927888e9a0e7c58e0cd87e843cf69f} \\
 & \texttt{24249df53eb34b92a68ce72f3c1ce8c7} \\
Joined semantics & \texttt{f4f9e625dded0c0ee712012f14cd7ea3} \\
 & \texttt{ad4f896ec991c2389139412dc2d584dd} \\
Prompt equivalence & \texttt{cf36ed363351ffed8990c9f2ff14f095} \\
 & \texttt{5b6ad3eca08afbcdca6b7c42a62efb98} \\
\end{tabular}}

The strict release identity binds the root and public manifests, all four
public and four private shards, the complete file inventory, and
\texttt{SHA256SUMS}; the prompt-equivalence identity binds the hash of every
model-facing prompt under its semantic join.  Same-seed regeneration is
byte-identical.  Every model run has its own run identifier, and no failed or
restarted run is silently overwritten.

\paragraph{Data statement.}

All linguistic forms, programs, meanings, labels, supports, and witnesses are
generated.  The artifact contains no people, personal data, speaker labels,
human annotations, or copyrighted corpus passages.  Synthetic strings reduce
direct textual contamination but not familiarity with logic-puzzle structure.

\section{Capability Map of Neighbouring Resources}
\label{app:capability-map}

\begin{figure*}[t]
  \centering
  \includegraphics[width=\textwidth]{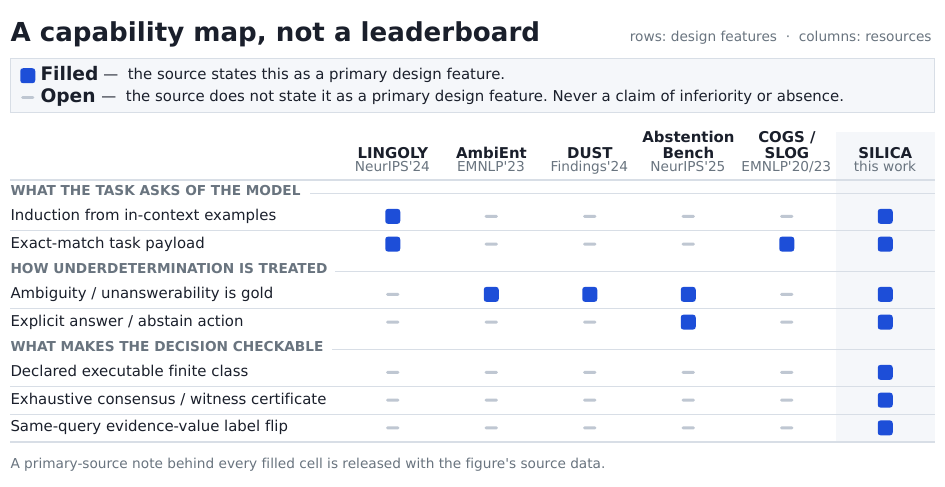}
  \caption{\textbf{A capability map, not a leaderboard.}
  Filled cells mark features explicitly stated as primary design features in
  the corresponding source; open cells mean only ``not stated as a primary
  design feature,'' never inferiority or total absence.  The columns cover
  in-context linguistic puzzles \citep{bean2024lingoly}, ambiguity and
  underspecification \citep{liu2023ambient,wildenburg2024dust}, explicit
  abstention \citep{kirichenko2025abstentionbench}, and compositional semantic
  parsing \citep{kim2020cogs,li2023slog}.  The source basis for every filled
  cell is released with the figure's source data.}
  \label{fig:capability-map}
\end{figure*}

\section{Decision Structure of the Registry}
\label{app:guard-structure}

The structural unit is the published Boolean program, not the semantic template.
Each of the 192 evaluation templates is instantiated as eight lexical--grammar
routes, and every route publishes its own program, its own revealed feature
triple, and its own pair of query-facing coordinates, so the registry holds
\textbf{1,536 distinct programs} over the 12,288 rows, eight rows each.
Our audit recovers each program's
decision function \emph{behaviourally}---by enumerating the declared universe and
asking which query-facing coordinate is constant over the survivors---so the
census does not depend on how a program happens to be spelled.

\paragraph{How many structurally different guards.}
The 1,536 programs realise \textbf{86 distinct three-bit decision functions} of
the 256 possible: 32 occur in the majority-guard family, 31 in parity-guard, 24
in selector-multiplexer, and 14 in equality-chain.  Those counts sum above 86
because the families are different encodings of one guarded-switch principle and
do realise some of the same decision functions, which is the honest reading of
the four-family design.  No guard is unique to one program, and the most frequent
covers 53 of 1,536 (3.5\%), so memorising one guard transfers to almost nothing
even before the held-out-template split.

\paragraph{What the decision actually costs.}
The two procedures differ by about three orders of magnitude, so the $2^8$
framing should not be read as implying that the task is computationally deep.
The \emph{stated} enumerate-and-filter route costs 256
declared-universe assignments times the published program's operator count
(median 22, range 6--40), a median of \textbf{5,632 operator evaluations} per
row, before the sixteen survivor query evaluations.  The \emph{minimal
sufficient} decision---which query-facing coordinate the evidence pins---is a
Boolean function of the three revealed bits with minimal formula size
\textbf{3 (median), at most 4} over the released DSL's own basis; 32 of the 1,536
programs need none at all.  The gold decision is therefore at most four Boolean
operations on three bits the prompt prints explicitly, which is why we read the
silent panel's decision signal of $[-.010,+.017]$ as a failure to consult
identifiability rather than an arithmetic shortfall.

\paragraph{Where the decision information lives.}
A compact rule that is available only \emph{after} the program has been evaluated
on the evidence is the task, not a leak.  Two gradient-boosted probes
\citep{friedman2001greedy,pedregosa2011sklearn} under the
same grouped splits as the released controls make that checkable.  A probe given
the three revealed bits, the query role, the mechanism family, the response task
and the declared class and posterior sizes---every evidence- and query-side
feature except the guard value---scores balanced decision accuracy $.500$ under both the eight-fold
held-out-template and the four-fold held-out-family split.  The same probe
additionally given the guard value scores $1.000$ under both.  The guard value is
not a field of the input: obtaining it \emph{is} evaluating the published program
against the evidence, and a predictor that never performs that evaluation sits at
chance whatever else it is given.

\section{Prompt Ablations on Two Panel Checkpoints}
\label{app:prompt-ablations}

Four renderings of the same 1,536-row cohort were served from one Qwen2.5-14B
endpoint and, separately, one Qwen2.5-32B endpoint, under greedy decoding at
seed zero.  \texttt{verbatim} is the released
prompt; \texttt{scratchpad} keeps the user message byte-identical and replaces
only the ``reason silently'' instruction with a \texttt{<think>} block and a
4{,}096-token budget, scored on the final JSON alone; \texttt{lean} keeps the
system message and removes nine fields from the user message: the grammar
realization block, the constant markers and rules, the fixed lexicon, the
schema family, the adapter's protocol hash, task and version, and the
response-schema version.  This shortens the message by 25.3\% on average while
keeping the feature-value declaration, without which the task is undecidable;
\texttt{compact\_rule} keeps the user message byte-identical and replaces the
enumerate-$2^8$ instruction with the guarded-switch route.  Because it states
structure the released prompt leaves to be inferred (though Section~\ref{sec:stress}
publishes it), \texttt{compact\_rule} is an upper bound on what relieving the
stated procedure can buy, not a matched control.

These are exploratory and run outside the sealed evaluation stack, so
\texttt{verbatim} is a validity check on the harness rather than a result: it
reproduces the sealed panel row for each checkpoint to a maximum absolute
difference of $.023$ (14B) and $.0052$ (32B) across strict format validity, AR,
FCR, UR and BJA, which is also a transformers-to-vLLM replication of two panel
conditions.  \texttt{lean} is
gated on decidability: the prompt-only solver still scores 12{,}288/12{,}288 on
the reduced rendering, so the arm removes redundancy and not evidence.

\begin{table}[t]
\centering
\small
\caption{Prompt ablations on Qwen2.5-14B, 1,536 rows each.  AR moves by a
factor of $8.6$ across arms while the decision signal never leaves
$[-.007,+.004]$ and the all-eight endpoint stays at $0/192$.}
\label{tab:prompt-ablations}
\resizebox{\columnwidth}{!}{%
\begin{tabular}{@{}lrrrrrr@{}}
\toprule
Arm & Valid & AR & AR$-$FCR & CPA & BJA & Tok. \\
\midrule
\texttt{verbatim} & .990 & .466 & $+$.004 & .503 & .382 & 18 \\
\texttt{scratchpad} & .984 & .730 & $-$.007 & .496 & .307 & 697 \\
\texttt{lean} & 1.000 & .203 & $-$.005 & .513 & .448 & 18 \\
\texttt{compact\_rule} & .998 & .085 & $+$.003 & .446 & .477 & 13 \\
\bottomrule
\end{tabular}}
\end{table}

The scratchpad arm is the informative one.  Qwen2.5-14B writes a mean of 697
tokens of visible working per row, closes the block on 1{,}510/1{,}536 rows, and keeps
$.984$ strict format validity, yet its decision signal is $-.007$ against $+.004$
for its verbatim twin on the same endpoint.  What the trace buys is a higher
answer rate, which \emph{lowers} balanced strict joint accuracy from $.382$ to
$.307$ by moving the policy toward the always-\texttt{ANSWER} constant.  Together
with the factor-isolation ladder this gives three points rather than one: a trace
moves a reasoning-trained 235B, does not move a reasoning-trained 1.7B, and does
not move a non-reasoning-trained 14B given a scratchpad and a budget.

Qwen2.5-32B, the panel's near-always-\texttt{ABSTAIN} checkpoint, gives the same
picture from the low end of the answer-rate range.  Across \texttt{verbatim},
\texttt{scratchpad}, \texttt{lean} and \texttt{compact\_rule} its AR is $.025$,
$.845$, $.146$ and $.010$, while its decision signal is $+.003$, $+.009$, $+.013$
and $+.001$ and the all-eight endpoint stays at $0/192$.  Its scratchpad arm
writes a mean of 832 tokens per row and closes the block on 1{,}502/1{,}536 rows,
and again only raises AR, lowering balanced strict joint accuracy from $.495$ to
$.264$.

\section{The Derivation, Step by Step}
\label{app:deduction}

\begin{figure*}[t]
  \centering
  \includegraphics[width=\textwidth]{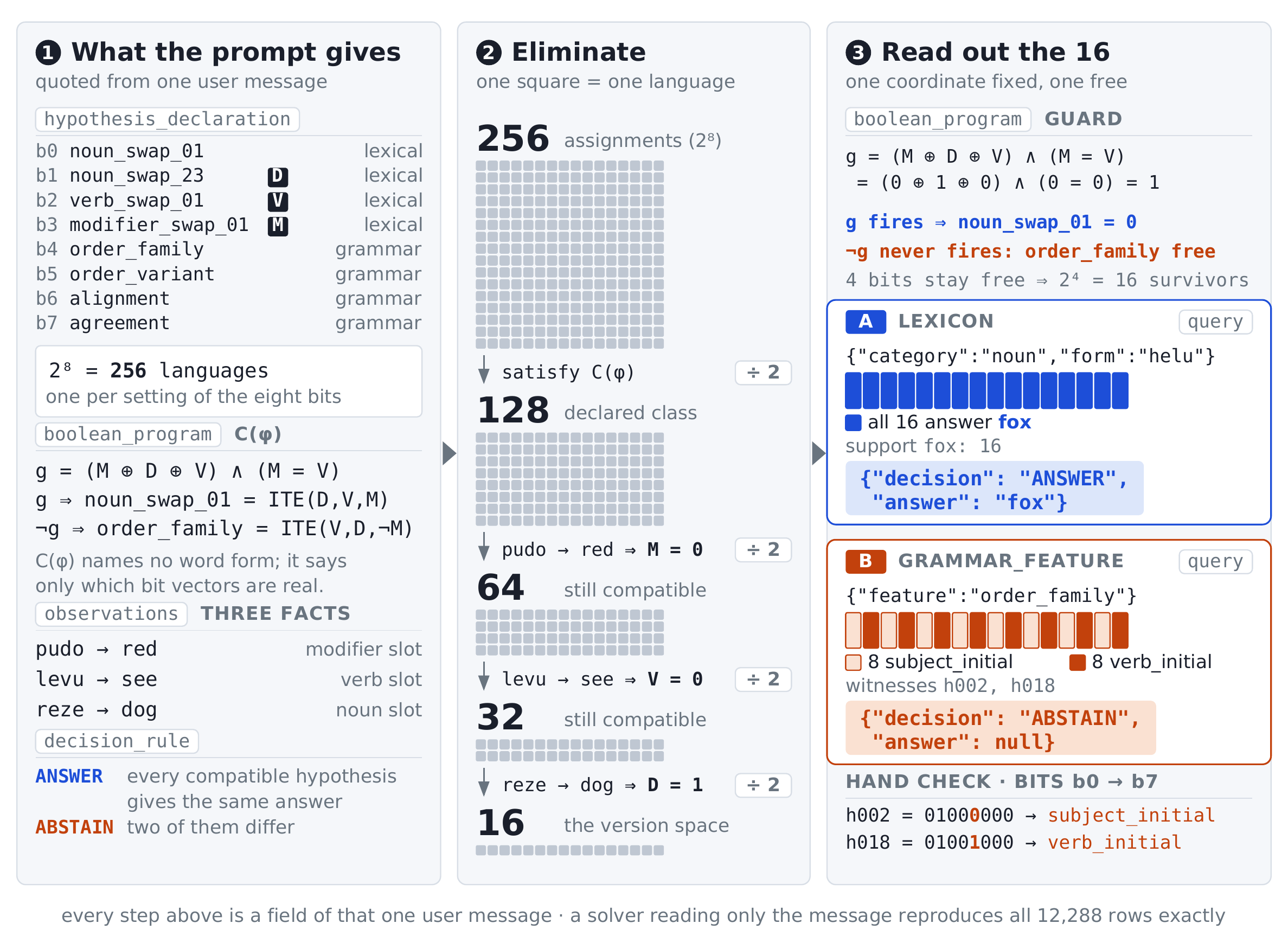}
  \caption{\textbf{One item, worked end to end.}
  \emph{1.} The prompt fields that carry the problem, quoted from a single user
  message: eight binary features, the public program $C_\phi$, three dictionary
  facts, and the \texttt{ANSWER}/\texttt{ABSTAIN} rule.  $C_\phi$ names no word
  form; it constrains only which bit vectors are real.
  \emph{2.} Elimination, one square per surviving language: 256 assignments,
  128 satisfying the program, then $64\rightarrow32\rightarrow16$ as each fact
  pins one feature bit.
  \emph{3.} Readout over those same 16.  The guard fires, so
  \texttt{noun\_swap\_01} is constant and the lexical query is unanimous
  (\textsc{fox}, 16/16, \texttt{ANSWER}); its negation never fires, so
  \texttt{order\_family} stays free and splits 8/8, forcing \texttt{ABSTAIN}
  with witnesses \texttt{h002} and \texttt{h018} shown as explicit bit vectors.
  Support counts and witnesses are evaluator-only and never appear in the
  prompt.}
  \label{fig:deduction}
\end{figure*}

Figure~\ref{fig:deduction} walks the complete derivation for the released parity
item of Figure~\ref{fig:main}, in the order a solver must perform it.  Panel~1
quotes the four fields of the single user message that carry the whole problem:
the eight-bit feature inventory, the public Boolean program, the three
dictionary facts, and the decision rule.  Panel~2 performs the elimination:
$2^8=256$ assignments, exactly half satisfying the program, then one halving per
observed fact down to the 16-member version space.  Panel~3 reads the two
complementary queries off that shared set, including the guard evaluation that
fixes one query-facing coordinate while leaving the other free, and a hand check
of the two disagreeing witnesses as explicit bit vectors.

Every quantity in the figure is recomputed from the sealed artifact rather than
transcribed, and the elimination counts are measured rather than assumed: each
of the three facts halves the surviving set exactly.  The construction guarantees
this: every one of the eight observed-bit states leaves exactly 16 of the 128
declared languages, and the halving holds on all 12,288 rows.  The figure
also states the sufficiency result of Section~\ref{sec:results}: because every
step shown is a field of that one message, a solver reading only the message
reproduces all 12,288 rows exactly.

\subsection{The exact instruction}
\label{app:instruction}

The system message is identical for every scored row and states the procedure
outright.  It is reproduced here verbatim:

\begin{quote}\small\ttfamily\noindent
You are solving the SILICA-Hard posterior-matched paired-query identifiability
reasoning benchmark. The supplied declaration defines an exact finite hypothesis
class: enumerate all Boolean feature assignments, retain only assignments
accepted by every declared Boolean-program constraint, and then retain only
assignments consistent with every observed form/meaning pair. Evaluate the query
under every surviving assignment. Return ANSWER only if all survivors produce
one identical canonical answer; otherwise return ABSTAIN. Reason silently. Emit
exactly one JSON object satisfying canonical\_response\_schema, with no
markdown, rationale, confidence, or additional keys.
\end{quote}
\noindent The benchmark name inside this message is the one the evaluated models
saw; like every other prompt byte, it is kept unchanged in the release.

\noindent
The user message repeats the same contract in structured form: the
\texttt{exact\_enumeration} field names the universe (``all $2^8$ assignments in
\texttt{feature\_variables\_lsb\_to\_msb} order'') and the filter, and the
\texttt{decision\_rule} field states both branches.  No worked example, chain of
thought, or few-shot demonstration is supplied, and the schema admits no key
beyond \texttt{decision} and \texttt{answer}.  Three properties of this
instruction matter for interpreting the panel.  The algorithm is \emph{given},
so the task is execution rather than discovery.  The reasoning is required to be
\emph{silent}, so a model must carry the enumeration through one forward pass.
And the contract is \emph{identical} across the two members of every same-state
pair, so a model that reads the query cannot be disadvantaged by the wording.

\section{The Solution, Object by Object}
\label{app:walkthrough}

\begin{figure*}[tp]
  \centering
  \includegraphics[width=\textwidth]{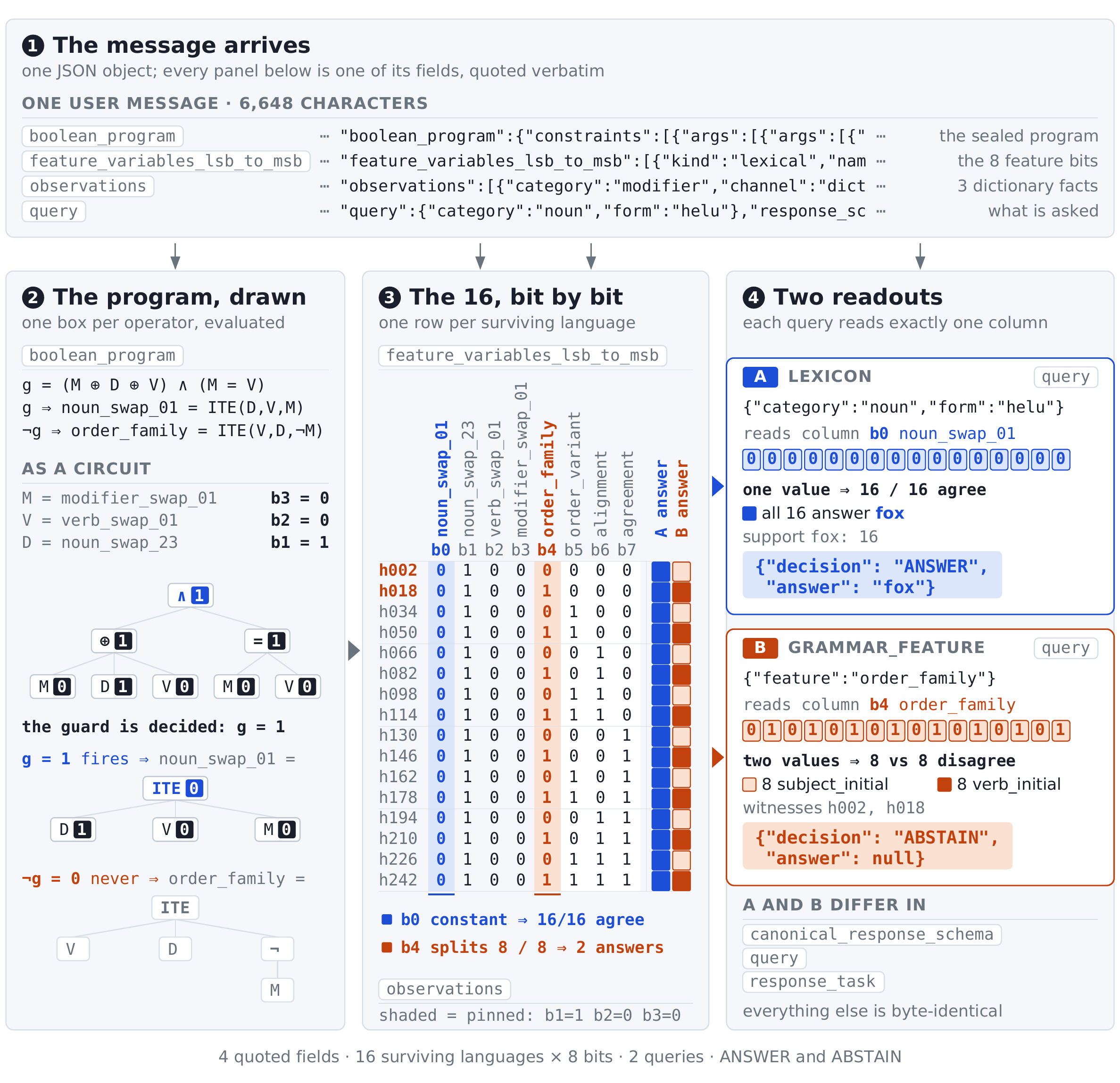}
  \caption{\textbf{One item, object by object.}
  \emph{1.} The rendered user message, quoted verbatim at each of the four keys
  that carry the problem.
  \emph{2.} The sealed Boolean program drawn as its own syntax tree, one box per
  operator, each carrying the bit it evaluates to under the three facts.  The
  guard is decided, so \texttt{noun\_swap\_01} is assigned and
  \texttt{order\_family} is never touched.
  \emph{3.} The 16 survivors as literal bit vectors over the 8 declared
  features.  Shaded columns are pinned by the observations; the blue column is
  constant, the orange column splits 8/8, and the two narrow strips on the right
  give each row's canonical answer to each query.
  \emph{4.} Each query lifts out its own column: one value gives \textsc{fox},
  16/16, \texttt{ANSWER}; two values give \texttt{ABSTAIN} with witnesses
  \texttt{h002} and \texttt{h018}.  The two prompts differ only in the three
  listed keys.  Support counts and witnesses are evaluator-only and never appear
  in the prompt.}
  \label{fig:walkthrough}
\end{figure*}

Figure~\ref{fig:walkthrough} shows what a solver literally holds at each step of
the same frozen pair, where Figure~\ref{fig:deduction} shows how the counts
collapse.  Panel~1 quotes the rendered user message itself: four contiguous
slices of the one 6,648-character JSON object, each tagged with the key it is
taken at, so nothing downstream is introduced from outside that message.
Panel~2 draws the public Boolean program as the abstract syntax tree the release
seals rather than as a string, with the bit the clean-room evaluator assigns to
every node; because the three dictionary facts fix every leaf, the guard is
decided and exactly one branch of the program fires.  Panel~3 is the version
space written out cell by cell: 16 rows, one per surviving language, by the 8
declared feature bits, with the three observation-pinned columns shaded and the
two queried columns inked.  Panel~4 lifts each queried column out of the table
and reads the scored JSON off it.

The point of the middle panel is that the paper's central claim is checkable by
eye.  Column \texttt{b0} is 0 in all sixteen rows, so the lexical query has one
answer and the gold action is \texttt{ANSWER}; column \texttt{b4} carries eight
zeros and eight ones, so the grammar query has two, and the gold action is
\texttt{ABSTAIN}.  Neither queried column is one of the observed columns: the
constant one is fixed by the program given the facts, and the split one is never
constrained at all.  Every cell, identifier, count, and payload in the figure is
recomputed from the sealed artifact and checked against the version-space and
support certificates before the figure is allowed to draw; the queried column
itself is \emph{measured}, by toggling each feature bit over the whole declared
universe and keeping the ones that move the canonical answer, rather than
matched by name.

\section{Exact Sealed Panel Numbers}
\label{app:model-panel}

Figure~\ref{fig:model-panel} plots the panel; this table prints it exactly.
Every value is emitted by the collector from the provenance-verified ledger,
so the two cannot disagree.

\providecommand{\ModelPanelRows}{}
\renewcommand{\ModelPanelRows}{%
Qwen2.5-0.5B$^{\dagger}$ & .423 & .426 & .418 & $+$.008 & .001 & .471 & .214 & .101 [.090,\,.112] & .000 \\
Qwen2.5-1.5B$^{\dagger}$ & .030 & .026 & .034 & $-$.008 & .000 & .450 & .013 & .006 [.003,\,.009] & .000 \\
SmolLM2-1.7B$^{\dagger}$ & .730 & .738 & .721 & $+$.017 & .000 & .429 & .369 & .158 [.146,\,.170] & .000 \\
Qwen2.5-3B & 1.000 & .995 & .997 & $-$.003 & .003 & .497 & .499 & .249 [.244,\,.253] & .000 \\
Phi-3.5-mini$^{\dagger}$ & .000 & .000 & .000 & $+$.000 & .000 & --- & .000 & .000 [.000,\,.000] & .000 \\
Qwen2.5-7B & 1.000 & 1.000 & 1.000 & $+$.000 & .000 & .497 & .500 & .249 [.244,\,.253] & .000 \\
Qwen2.5-14B & .983 & .443 & .453 & $-$.010 & .531 & .497 & .487 & .376 [.360,\,.391] & .000 \\
Qwen2.5-32B & 1.000 & .030 & .025 & $+$.005 & .975 & .522 & .503 & .495 [.491,\,.499] & .000 \\
\midrule
always-\texttt{ANSWER} & 1.000 & 1.000 & 1.000 & $+$.000 & .000 & .500 & .500 & .250 & .000 \\
always-\texttt{ABSTAIN} & 1.000 & .000 & .000 & $+$.000 & 1.000 & --- & .500 & .500 & .000 \\
exact oracle & 1.000 & 1.000 & .000 & $+$1.000 & 1.000 & 1.000 & 1.000 & 1.000 & 1.000 \\
}

\begin{table*}[t]
\centering
\small
\caption{Sealed common-cohort panel: 1,536 rows (768 answerable, 768
underdetermined) over 192 semantic templates, one complete route each.  Every
value is emitted from the provenance-verified ledger; all eight conditions have
request success 1.000.  $\dagger$ marks conditions below the pre-output-frozen
0.95 format gate, reported but excluded from the hardness claim.
$\langle 8\rangle$ is the all-eight strict paired endpoint over 192 route
groups.  The two constant policies and the exact oracle are construction
references, not competitors.}
\label{tab:model-panel}
\begin{tabular}{@{}lrrrrrrrlr@{}}
\toprule
Condition & Valid & AR & FCR & AR$-$FCR & UR & CPA & BDA & BJA [95\% CI] & $\langle 8\rangle$ \\
\midrule
\ModelPanelRows
\bottomrule
\end{tabular}
\end{table*}

\section{Scope and Caveats}
\label{app:caveats}

This appendix details each point summarized in the Limitations section.

\paragraph{Scope of the instrument.}
\silica{} uses small synthetic inventories and simplified grammar.  It does not
model language acquisition, documentation, translation quality, or typological
frequency, and identifiability is exact only within each declared 128-member
class drawn from a 256-language inventory---not among all plausible human
analyses.  The shortcut audits cover the families we specified, not every
heuristic a future model might find.

\paragraph{The silent envelope is a confound, and we price it.}
The envelope is deliberately restrictive: the system message states the procedure
and then requires the model to reason \emph{silently}, the schema admits no key
beyond \texttt{decision} and \texttt{answer}, and 384 new tokens are allowed.  The
collapse is not a truncation or formatting artifact for the four conditions at
$\geq.98$ format validity, but it does measure single-pass execution, and that
constraint is not innocuous.  Lifting it on one checkpoint is worth $+.436$ of
decision signal and $11/192$ complete route groups at 235B, and $+.030$ and
$0/192$ at 1.7B.  Every claim in Section~\ref{sec:results} about the panel is
therefore a claim about the silent single-pass envelope, not about what these
models can do given room to derive; the trace-permitted conditions are reported
as a separate stratum and never folded into the panel.  The envelope explains
part of the collapse and not all of it, since permission plus a
$16{,}384$-token budget leaves 1.7B on the frontier, and a visible scratchpad
leaves Qwen2.5-14B and 32B there too (Table~\ref{tab:matched-controls}).  The
decision axis in particular is computationally shallow:
Appendix~\ref{app:guard-structure} shows the gold decision is a Boolean function
of three printed bits with minimal formula size at most four.

\paragraph{Breadth of the model evidence.}
All four conditions that pass the format gate are Qwen2.5 checkpoints, two of
which sit near a constant policy, so the format-valid evidence comes from one
model family.  The trace-permitted and decoding conditions add Qwen3 and
Qwen3.5 checkpoints from 0.8B to 235B, but they are exploratory controls, each
varying one factor on one checkpoint, not a scaling study.

\paragraph{Response validity is a capability we measure, not a nuisance.}
Four of the eight panel conditions fall below the $.95$ format gate, and in the
trace-permitted stratum the ability to \emph{close} a derivation inside the
budget varies from $0.0\%$ non-termination at 235B to $28.8\%$ at Qwen3.5-0.8B.
Reporting these as excluded rows understates the point: at small scale, the
constraint that binds is producing a well-formed answer at all, and no statement
about identifiability can be extracted from such a condition in either direction.

\paragraph{The single-EOS stop contract.}
The local panel terminates generation at a single declared EOS identifier, which
invalidates models whose turn boundary is a different token.  This left the
prespecified Phi-3.5 condition unevaluable; the replacement condition with a
multi-token stop contract is described in Appendix~\ref{app:protocol}.

\paragraph{Two implicit conventions.}
The declaration binds each lexical axis to its feature bit only through matching
name stems and list order, and reads \texttt{identity}/\texttt{swapped}
against the declared pair order.  Our prompt-only solver recovers both
conventions and scores perfectly, but they are implicit, and a future release
should declare the binding outright.

\paragraph{What the result does and does not generalize to.}
The registry is broad inside its phenomenon and narrow outside it.  It holds
1,536 distinct published programs over 192 templates, realising 86 distinct
three-bit decision functions, and the trace-permitted escape appears in all
four mechanism families ($+.354$ to $+.479$ decision signal) and across the
lexical and grammar query axes (.669 versus .672 balanced strict joint accuracy).
That is dense coverage of \emph{one} query-conditioned identifiability switch in
a closed, declared lexicon--grammar class.  It is not evidence about open-domain
underspecification, tool or API discovery, or real low-resource translation, and
we make no claim about those.  A resource that varied the mechanism rather than
its encoding is the main thing a future version should add.

\paragraph{Statistical scope.}
Greedy decoding has no sampling variance, template-cluster intervals condition
on one finite registry and omit benchmark-design uncertainty, and hardness is
claimed only for complete, provenance-valid runs meeting the pre-output-frozen
coverage and formatting gates.

\end{document}